\documentclass{article}

\usepackage[margin=1in]{geometry} 
\usepackage{lipsum}
\usepackage{enumitem}

\usepackage[utf8]{inputenc} 
\usepackage[T1]{fontenc}    
\usepackage{url}            
\usepackage{booktabs}       
\usepackage{amsfonts}       
\usepackage{nicefrac}       
\usepackage{microtype}      
\usepackage{graphicx}
\usepackage{natbib}
\usepackage{doi}
\usepackage{float}
\usepackage{amsmath}
\usepackage{amsthm}
\usepackage{multirow}
\usepackage{algorithm}
\usepackage{algpseudocode}
\usepackage{xcolor}
\usepackage{hyperref}
\hypersetup{
    colorlinks,
    linkcolor={red!50!black},
    citecolor={blue!50!black},
    urlcolor={blue!80!black}
}

\setcitestyle{authoryear, open={(},close={)}}

\newcommand{\beginsupplement}{%
        \setcounter{section}{0}
        \renewcommand{\thesection}{S\arabic{section}}%
        \renewcommand{\theHsection}{S\arabic{section}}
        \setcounter{table}{0}
        \renewcommand{\thetable}{S\arabic{table}}%
        \renewcommand{\theHtable}{S\arabic{table}}
        \setcounter{figure}{0}
        \renewcommand{\thefigure}{S\arabic{figure}}%
        \renewcommand{\theHfigure}{S\arabic{figure}}
     }

\title{\textbf{Toeplitz MLP Mixers are Low Complexity, Information-Rich Sequence Models}}

    \author{\small Benjamin L. Badger \\
    \small IBM\\
    \texttt{\small ben.badger@ibm.com}
    \and
    \small Ethan Roland \\
    \small AE Studio\\
    \texttt{\small ethan@ae.studio}
    }

    \date{}
    
\begin{document}
    \maketitle
    \footnotetext{The authors would like to thank IBM and AE Studio for support during the writing of this paper. Code may be found on \url{https://github.com/ethan-w-roland/ToeplitzMixers}}

\begin{abstract}\normalfont{
    Transformer-based large language models are in some respects limited by the quadratic time and space computational complexity of attention. We introduce the Toeplitz MLP Mixer (TMM), a transformer-like architecture that swaps attention for triangular-masked Toeplitz matrix multiplication over the sequence dimension resulting in $\mathcal{O} (dn \log n)$ time and $\mathcal O(dn)$ space complexity during training and $\mathcal O(dn)$ time and space at inference prefill. Despite the lack of sophisticated input modulation or state maintenance present in other sub-quadratic architectures, TMMs yield greater training efficiency in terms of loss achieved per compute and device memory. We demonstrate that TMMs are capable of retaining more input information resulting in improved copying ability, which we argue results from a lack of architectural biases. Consistent with higher input information retention, TMMs exhibit superior information retrieval and in-context learning benchmark accuracy compared to comparable architectures. We conclude with an analysis from the perspective of operator index theory and show that, counterintuitively, trained Toeplitz layers of causal non-invertible models are more likely to be invertible or nearly so than models that are actually invertible over their inputs.}
\end{abstract}

\section{Introduction}
    
    Shortly after the Transformer \citep{vaswani2023attentionneed} was introduced, attempts were made to reduce this model's $\mathcal O(n^2)$ time complexity without sacrificing causal (next token prediction) language training efficiency or the emergent multitask learning characteristic of this architecture \citep{radford2019language}. It is clear that this is a difficult proposition: any attempt to reduce the quadratic complexity of self-attention is equivalent to imposing certain restrictions on how information from each input token may be accessed, but nearly arbitrary token information access is a hallmark of many tasks that language models are used for today. Among the myriad of tasks Transformer-based LLMs are applied to that require near-arbitrary information processing, especially notable are in-context learning for tool calling agents, large-context information retrieval, chain-of-thought reasoning, document summarization, and many forms of code generation. From this perspective it is apparent that any restriction placed on which previous tokens can be accessed to predict any next token must be done carefully.
    
    What may be thought of the most severe restriction is to access only a single element or more commonly a single vector, a state that represents all previous tokens, and such models are of linear time and constant space complexity at inference. Attempts to impose linear complexity on attention via linearizing the dot-product softmax operation were introduced by \citep{shen2019efficient} and \citep{katharopoulos2020}, but have been found to be relatively inefficient to train \citep{poli2023hyenahierarchylargerconvolutional} and numerically unstable for large-scale causal modeling. Success has been found with a reformulation of linear-complexity models in the form of state-spaces, frameworks where the focus in on a moving average of a global context vector (similar to the hidden layer of an RNN) with typified update patterns. A particularly notable state space architecture for language modeling is the Mamba architecture introduced by \citep{gu2024mamba} and improved upon in \citep{dao2024transformersssmsgeneralizedmodels}; these models effectively overcome the training inefficiences and numerical instabilities plaguing linear attention by introducing selectivity into the state space formulation via performing token-dependent transformations. There are drawbacks to this approach, however: investigators have found that Mamba models exhibit substantially worse copying ability \citep{jelassi2024repeatmetransformersbetter} and in-context learning \citep{waleffe2024empiricalstudymambabasedlanguage} compared to Transformers that achieved similar training losses. The advantages and disadvantages of these these linear-complexity models may be thought of as being intimately tied to the use of constant memory state spaces, which allow for greater inference speed and parallelism but also place severe limitations onto how much token input information these models can store, particularly for out-of-distribution inputs.

    Less severe restrictions on token operations may be imposed by restricting access to several elements, or by structuring the pattern of access itself. In the latter case, the benefit of doing so is typically lower computational complexity for forward passes in which many tokens are present, which is the usual scenario for training as well as first token generation from a large prompt or inference prefill (the computation that occurs to generate the first next token). This is the approach of state space models such as H3 \citep{fu2023hungryhungryhipposlanguage}, Hyena/H4 \citep{poli2023hyenahierarchylargerconvolutional}, and Monarch Mixers \citep{fu2023monarchmixersimplesubquadratic} which all exhibit $\mathcal O(n \log n)$ time complexity during training and inference prefill.
    
    We investigate a simple architecture that imposes a structured access pattern without restricting the number of tokens accessed: a Masked Mixer \citep{badger2025maskedmixerslanguagegeneration}-derived architecture where the token mixing weights are Toeplitz matrices in which all values on each diagonal are identical, which we call the Toeplitz MLP Mixer (TMM). Toeplitz matrices may be thought of as specifying Fourier Transform (DFT to be precise) coefficients, so this restriction allows for Fast Fourier Transform-based computation that is $\mathcal O(n \log n)$ time and $\mathcal O(n)$ space complexity. We find that this architecture exhibits greater training efficiency terms of loss achieved per billions of tokens in real wall-clock terms relative to other linear-complexity models in our experiments. We show that when trained for causal autoregressive generation, TMMs retain more input information and furthermore have a higher informational capacity than other linear-complexity models, and that this effectively closes the copy ability gap with quadratic-complexity models. This information retention ability is found to be reflected in benchmark accuracy, suggesting it is functionally beneficial to autoregressive generation. We offer a perspective into what linear complexity language modeling requires in practical and theoretical considerations, and present the TMM as an example of a model that provides beneficial characteristics in these terms.

\section{Our Contribution}
    Our work is based on the following assumption:

    \vspace{0.15cm}
    \noindent{\textit{Reducing the causal language modeling computational complexity restricts the pattern of information access during next token prediction, motivating an investigation of both training efficiency and informational processing in these models.}}
    \vspace{0.15cm}

    \noindent The fundamental insight of this work is as follows:
    
    \vspace{0.15cm}
    \noindent{\textit{Parameterizing token mixing weights as trainable Toeplitz matrices allows for lower computational complexity with relatively small decreases in training efficiency and information retention.}}
    \vspace{0.15cm}

    \noindent{We make the following claims:}

    \vspace{0.15cm}
    \begin{enumerate}[nosep]
    \item Training efficiency: Under constant compute conditions, TMMs reach similar or lower evaluation loss compared to other sub-quadratic models.
    \item Information preservation: TMMs retain more input information as shown by information extraction, model inversion, copy experiments, and information retrieval benchmarks compared to other sub-quadratic models.
    \end{enumerate}
    
\section{Background and Method}

    The TMM is essentially a Masked Mixer \citep{badger2025maskedmixerslanguagegeneration} with token mixing operations parameterized by Toeplitz matrices in which all values on a given diagonal are identical, and thus the entire matrix may be represented by one vector (the first row and column). When applied to sequence modeling, Toeplitz matrix parameterization also gives the desirable property that weights of previous tokens are invariant on a relative position basis, or in other words no matter the current token index, the token at $n$ positions before the current token will always have the same weight. The Masked Mixer is an MLP mixer (Transformer-like models that swap attention for MLPs introduced by \citep{melaskyriazi2021needattentionstackfeedforward, tolstikhin2021mlpmixerallmlparchitecturevision}) adapted for causal sequence modeling. We enforce causality by triangular masking of the materialized weight matrix, which we denote below by $T'$, for both training and inference. In our implementation, we left-multiply activations by Toeplitz matrices and therefore $T'$ is upper triangular for (left-right) causality. 
    
    More formally, in the token mixing layer given input $X$ with hidden dimension $D$ and $n_{ctx}=N$, $X\in \Bbb R^{D \times N}$, we multiply by $T' \in \Bbb R^{N \times N}$ and broadcast-add a per-token-index bias $B \in \Bbb R^{N}$ to receive $Y \in \Bbb R^{D \times N}$ as shown in Equation \ref{eq1}. 

    \begin{equation}
    Y = XT' + B
    \label{eq1}
    \end{equation}

    An $\mathcal{O}(dn \log n)$ time and $\mathcal O(dn)$ space implementation of Equation \ref{eq1} is as follows: as every Toeplitz matrix may be embedding in a circulant matrix $C$, we represent $T'$ by the first column of $C$ which we denote $T_c$ (here the causal mask is simply zeros for indices that are triangular masked in the materialized matrix) and for each column vector $x \in X$, zero padded to match the length of $T_c$, and with Hadamard (element-wise) multiplication denoted as $\odot$, we can calculate $XT'$ using the Fast Fourier Transform $F$ as shown in Equation \ref{eq4}. 
    
    More precisely, we adapt the usual FFT-based fast Toeplitz matrix right-multiplication computation to left-multiplication by using the property $Y = TX = (X^T T^T)^T$, denoting the circulant vector form of $T'$ as $T_c$ and the padded $X$ as $X_p$. First we obtain the FFT of each row (sequence dimension) of $X_p^T$ and the FFT of the first column vector of the circulant embedding of the transpose of $T_c$. We then perform Hadamard multiplication of the resultant vectors, obtain the inverse FFT of this vector, and finally concatenate all vectors and transpose to obtain $Y$. As any module using $XT'$ may be expressed using (\ref{eq4}), we denote this multiplication as simply $XT'$ going forward. Based on the official SciPy \citep{virtanen2020scipy} implementation for fast Toeplitz matrix-matrix multiplication via FFTs, our Pytorch implementation of (\ref{eq4}) is compatible with many useful features such as mixed precision training and inference, distributed launch, and Pytorch compilation.

    \begin{equation}
    Y = XT' + B= F^{-1} \left( F(T_c^T) \odot F(X_p^T) \right)^T + B
    \label{eq4}
    \end{equation}

    We also implement a multi-headed Toeplitz token mixing layer where each head receives a trainable input projection $I \in \Bbb R^{D \times d_h}$ indexed by the number of heads $h$ with head dimension $d_h$ by applying independent Toeplitz matrices to each head, and concatenating (denoted by $\circ$) the outputs of each head together before multiplying by an output projection $P \Bbb \in R^{d_h*h \times D}$ as shown in Equation \ref{eq2}. Equivalently (and with somewhat more efficiency) we can project into one input matrix $S$ by multiplying $X$ by $I \in \Bbb R^{D \times d_h * h}$, splitting $S$ in the hidden dimension into $h$ matrices, right-multiplying those by Toeplitz matrices, and proceeding with concatenation and output projection.

    \begin{equation}
        Y = P\bigg ((I_0X)T'_0 + B_0 \circ (I_1X)T'_1 + B_1   \circ \cdots \circ I_{h-1}(X)T'_{h-1} + B_{h-1} \bigg )
        \label{eq2}
    \end{equation}

    For our models, we implement multi-headed Toeplitz layers with $D = h * d_h$ such that input and output projections are $I = P = \Bbb R^{D \times D}$.

    Inspired by the close relationship between Toeplitz matrices and convolutions, we also implement a token mixing layer in which the `kernel' value is greater than one: in this case, each token mixing layer mixes along both token (sequence) dimension as well as along a limited number of hidden dimension elements, the latter parameterized by the size of the kernel. We train separate Toeplitz layers for each kernel element, similarly to 1-dimensional convolutional filters, as shown in Equation \ref{eq3}. We zero pad $X$ to make $X \in \Bbb R^{d + k \times n}$ and use a stride of size 1.

    \begin{equation}
    Y = B + \sum_{i=0}^{k} X_{(i:d+i, \;:)}\;T'_i
    \label{eq3}
    \end{equation}

\begin{figure}[h]
        \centering
        \includegraphics[width=0.95\textwidth]{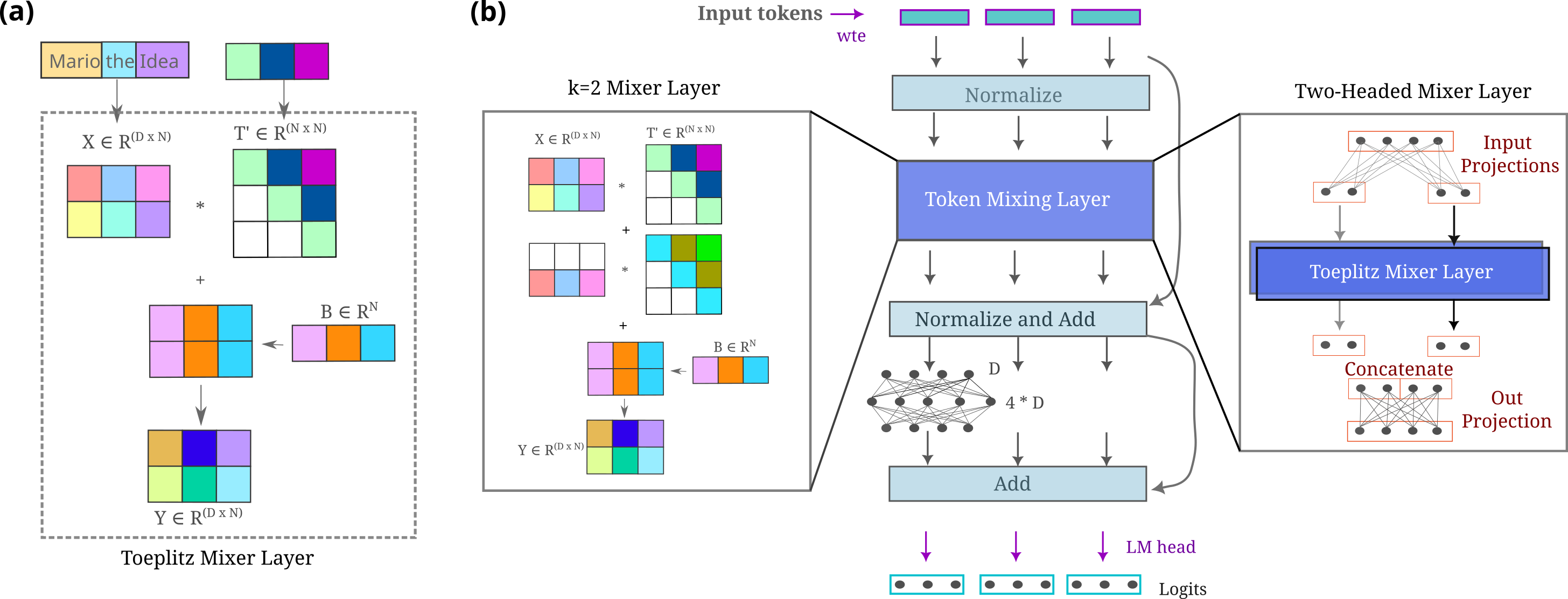}
        \caption{TMM architecture examples. (a) Illustration of a single Toeplitz mixer layer and (b) a TMM module with either a two-headed or k=2 Toeplitz layer. Models in this work are typically composed of 16 modules in sequence.}
        \label{fig2}
    \end{figure}

    For clarity, the Toeplitz matrix-based token mixing operation, multi-head configuration, and multi-kernel configuration is given in Figure \ref{fig2}.

\section{Computational Complexity}

    We elaborate on the computational complexity of TMMs and related models in this section. 

    \subsection{Training and Prefill: Without caching}
    
    Dot product self-attention in the Transformer \citep{vaswani2023attentionneed} exhibits $\mathcal O(n^2 * d)$ time complexity, which is evident by the multiplication of $QK^T$ with $Q \in \Bbb R^{n \times d_k}$ and $K^T \in \Bbb R^{d_k \times n}$ yielding an intermediate state $S \in \Bbb R^{n \times n}$ which is softmax-transformed and multiplied by $V \in \Bbb R^{n \times d_v}$. The formation of $S$ requires $n^2 * d_k$ operations, and the multiplication by $V$ requires $n^2 * d_v$ operations. 

    In pure matrix multiplication form, the token mixing of a causal $k=1$ TMM may be represented as the matrix-matrix multiplication of (lower-triangular masked) Toeplitz matrix $T \in \Bbb R^{n \times n}$ by input $X \in \Bbb R^{n \times d}$, with $\mathcal O(n^2 * d)$ time complexity. By embedding $T$ into the circulant matrix $A$ of size $2n$, the matrix-vector product of $Td_i$ can be computed via Fast Fourier Transforms via Equation \ref{eq4} with $\mathcal O(n \log_2 n)$ complexity, such that $XT$ exhibits $\mathcal O(n \log_2 n * d)$ time complexity.

    Space complexity for TMMs is $\mathcal O(n * d)$, which results from the size of the vector necessary to represent the first row and column of the Toeplitz matrix (which is of size $n$) and there are $d$ vectors which are multiplied via FFTs. This is lower than that for Transformers, which require $\mathcal O(n^2d)$ space, and avoids the quadratic bottleneck usually associated with masking as noted in \citep{fu2023monarchmixersimplesubquadratic}. 

    \subsection{Autoregressive Cached Inference}

    During autoregressive inference (assuming that $K, V$ rows are cached) Transformers are $\mathcal O(nd)$ time and space complexity for each new token generated as $QK^T$ multiplies $Q\in \Bbb R^{1 \times d_k}$, $K^T \in \Bbb R^{d_k \times n}$ such that this matrix-vector product is $\mathcal O(dn)$ creating $S \in \Bbb R^{1 \times n}$, and multiplication by $V \in \Bbb R^{n \times d_v}$ is again $\mathcal O(dn)$. TMMs exhibit the same complexities assuming that all rows except the current of $X$ are cached, as here the relevant multiplication is the last row of $T$, $T_{-1} \in \Bbb R^{1 \times n}$ and $X \in \Bbb R^{n \times d}$. 

\section{Related Work}
    The architectures presented in this work were inspired in part by the Masked Mixers introduced in \citep{badger2025maskedmixerslanguagegeneration}. A significant disadvantage of that architecture for long-context generation is the $\mathcal O(n^2)$ space complexity for $n$ tokens necessary to store the token mixing convolutions: the TMM reduces this to $\mathcal O(n)$ space as all Toeplitz matrices are representable by a single vector, and further allows for $\mathcal O(n \log n)$ rather than $\mathcal O(n^2)$ time complexity via FFT-based fast matrix multiplication. The substitution of arbitrary linear layers with Toeplitz-decomposed linear layers was found to have minimal adverse effects on small-scale vision modeling \citep{liu2022ludecompositiontoeplitzdecomposition}, making these transformations ideal candidates to replace Masked Mixer linear layers.
    
    The evolution of state space models from H3 \citep{fu2023hungryhungryhipposlanguage} to Hyena \citep{poli2023hyenahierarchylargerconvolutional} resulted in models that were increasingly efficient to train with $\mathcal O (n \log n)$ complexity at train time and inference prefill. The TMM is perhaps most similar to the Hyena model, which contain token mixing operations parameterized by layers of alternating scaling and Toeplitz matrix multiplication together with trainable trigonometric scaling functions and a windowed activation where more recent tokens receive higher weight. Our architecture departs from the Hyena in a number of ways: firstly as we use what is sometimes called `explicit' convolution parameterization where the model learns the weights of the convolution directly rather than `implicit' parameterization where it does not, secondly we do not add the recency and frequency biases present in that model, and thirdly TMMs contain only single sequential linear transformations for token mixing. We argue the use of explicit convolutional parameters is beneficial as it greatly reduces the number of activations formed during training, resulting in a smaller effective model.

    Another approach to reducing the quadratic complexity of attention to $\mathcal{O}(n \log n)$ is to use windowed local attention combined with a low-complexity global attention \citep{beltagy2020longformerlongdocumenttransformer} or else local attention and key hash-based bucketing \citep{kitaev2020reformerefficienttransformer}. 
    
    At an extreme, efforts have been made to reduce complexity all the way down to $\mathcal O(n)$ time and $\mathcal O(1)$ space at inference, although this class of model long predates Transformers (see augmented recurrent neural networks such as the LSTM \citep{hochreiter1997long}). The introduction of the linear-complexity selective state space architecture Mamba \citep{gu2024mamba} with sophisticated memory routing GPU kernels was a notable improvement on these architectures in the perspective of inference speed and computational complexity. Reformulations of that architecture, termed Mamba 2 \citep{dao2024transformersssmsgeneralizedmodels}, exceed the training efficiency of the original Mamba model and are therefore used for reference in this work


\section{Experimental Results}

    \subsection{Experimental Setup: Training}

    We train on a 10 billion token subset of an educational subset of the FineWeb called FineWeb-edu \citep{penedo2024finewebdatasetsdecantingweb}, a diverse dataset filtered from the Common Crawl and further selected for educational content. We implement models, custom loss functions, and special metrics using Pytorch \citep{paszke2019pytorchimperativestylehighperformance} and integrate these into the Huggingface \texttt{transformers} \citep{wolf-etal-2020-transformers} trainer utility. Datasets were typically pre-processed (tokenized, batched, and padded) before training and loaded on-demand as \texttt{datasets} \citep{lhoest-etal-2021-datasets} objects during training, which we find prevents latencies associated with slow disk reads or CPU bottlenecks due to fragmented storage or tokenization demands. Copy task datasets were generated on-demand from FineWeb inputs. All losses are reported for hold-out evaluation data unless otherwise noted.

    We train with peak learning rates of $\eta=5 * 10^{-4}$ for TMMs and Hyena models and $\eta=2 * 10^{-4}$ for Transformers and Mamba 2 (which we refer to simply as `Mamba') models with linear warm-up (over the first 500 samples) and learning rate decay over the (typically 200,000 step) training run, optimizing via AdamW \citep{loshchilov2019decoupledweightdecayregularization}. We note that we see little difference in loss upon changing learning rates for these models, likely due to our use of learning rate decay which has been observed to effectively negate loss differences resulting from different peak learning rates \citep{olmo20252olmo2furious}. TMMs, Transformers, and Mixers are $n_l=16, h=4$ unless otherwise noted, Mamba models are $n_l=16, h=8$ and Hyena models are order two with three features (similar to heads) and a filter order of 64 unless otherwise noted. We use Kaiming He normal initialization \citep{he2015delvingdeeprectifierssurpassing} at the start of training.

    Throughput and memory usage during training are given in Section \ref{throughputsection}: in general, inter-token transformations in TMMs require fewer activations which allows one to train models of twice the width $d_m$ than Transformers and Hyena models all else equal, or around quadruple the $d_m$ of Mamba models. We note that for our training context windows ($n_{ctx}\leq 1024$) matrix multiplication of activations by the materialized Toeplitz layer results in slightly higher throughput than FFT-based Toeplitz matrix multiplication as the former makes more efficient use of GPU tensor cores, but for longer sequences the reverse is the case. 
    
    \subsection{Training efficiency}

\begin{figure}[h]
        \centering
        \includegraphics[width=1\textwidth]{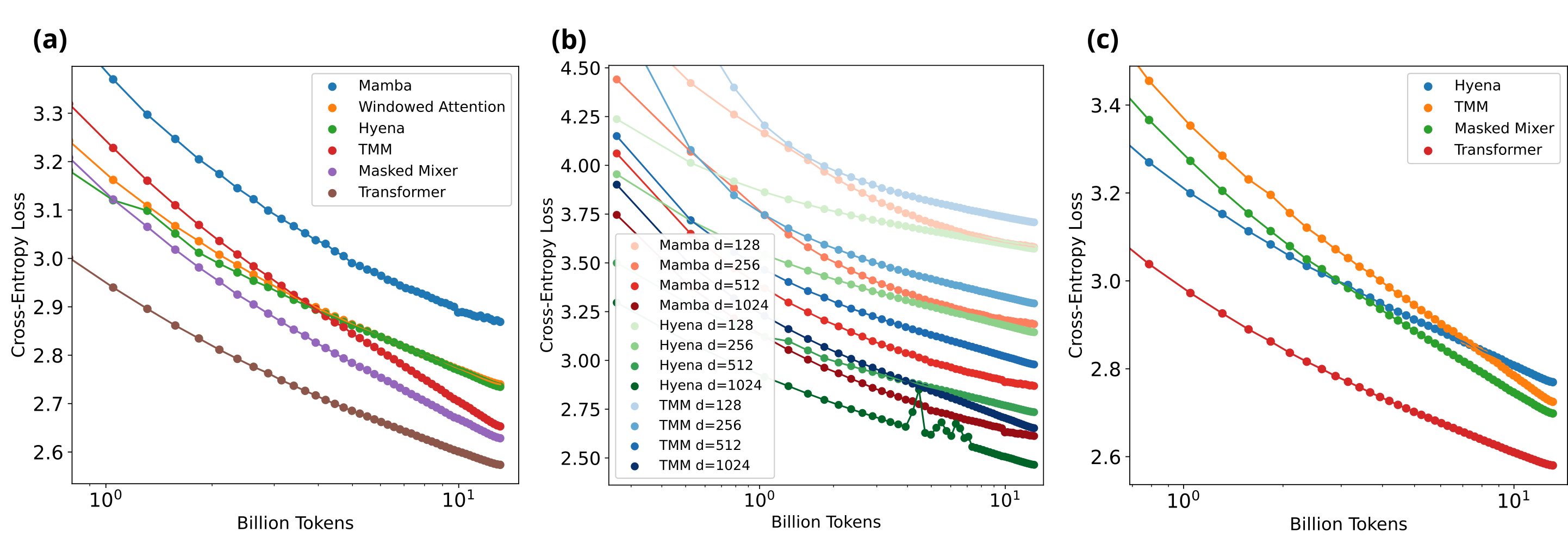}
        \caption{Training Efficiency on FineWeb. (a) Compute- and memory-matched examples of each architecture. (b) Model dimension scaling. (c) $n_{ctx}=1024$ compute-matched model scaling. (a) and (b) are $n_{ctx}=512$}
        \label{fig3}
    \end{figure}

    One of the primary challenges of architectures that restrict inter-token transformations in order to reduce computational complexity is relatively slow convergence during training, which we consider to be equivalent to the notion of inefficient training.  Slow convergence is a problem as large-scale language model training is a highly resource-intensive task, such that decreases in training efficiency result in models that become infeasible to train. Upon comparing TMM training efficiency to both quadratic (Transformer and Masked Mixer) and other sub-quadratic architectures (Hyena, windowed attention Transformers, and Mamba), we find that TMMs outperform compute- and memory-equivalent sub-quadratic but not quadratic models (Figure \ref{fig3}, and see Tables \ref{tables1} - \ref{tables4} for training throughput). We note that our implementation of causal Linear Transformers converge slowly relative to compute-matched TMMs and suffers from catastrophic numerical instabilities early in training (regardless of BF16/TF32 datatype use, layer renormalization, reformulation of causality training), so we do not investigate this architecture further (Figure \ref{figs0}).

    We also investigate the training efficiency of TMMs relative to models with restricted token information access via windowed (quadratic-complexity) attention in which each attention layer attends to a subset of input tokens, in our case the last $l$ tokens before the current token. As shown in Figures \ref{fig3} and \ref{figs3}, TMMs are more efficiently trainable than windowed attention even when $l$ is increased to equal half the context length, $l=n_{ctx}/2$.

    
    \subsection{Architectural Variations and Ablations}

    We investigated the training efficiency of TMMs with token mixing layers consisting of single Toeplitz layers as introduced in \citep{badger2025maskedmixerslanguagegeneration} versus two sequential layers separated by a nonlinearity as introduced in the original MLP mixer work \citep{melaskyriazi2021needattentionstackfeedforward, tolstikhin2021mlpmixerallmlparchitecturevision}. We find that two-layer TMMs exhibit lower training efficiency in terms of loss per training (Figure \ref{figs1}) and lower throughput, and do not investigate sequential multi-layer Toeplitz models further.

    Increasing the number of heads or the Toeplitz `kernel' size resulted in moderate increases in per-step FineWeb training efficiency to a point (four heads or k=4 kernel) but beyond a decrease in training efficiency despite increasing the number of inter-token parameters (Figure \ref{figs1}). Curiously increasing the head number but not kernel size results in superior asymptotic characteristics, with a relative increase in convergence rate later in training (Figure \ref{figs1}). After factoring in the difference in sample throughput during training (Tables \ref{tables2} and \ref{tables3}) the increased training efficiency of four-headed TMMs remains.

    We investigated whether or not trainable Toeplitz transformations and head projection transformations are necessary for efficient causal language model training, and as shown in Figure \ref{figs2} they are indeed: freezing Toeplitz matrices results in substantially lower FineWeb training efficiency regardless of whether or not head projections exist, although models with frozen Toeplitz matrices train more efficiently when head projects are present than when there are no head projections. 


    \subsection{Information Retention and Capacity} \label{informationsection}

\begin{figure}[h]
        \centering
        \includegraphics[width=0.85\textwidth]{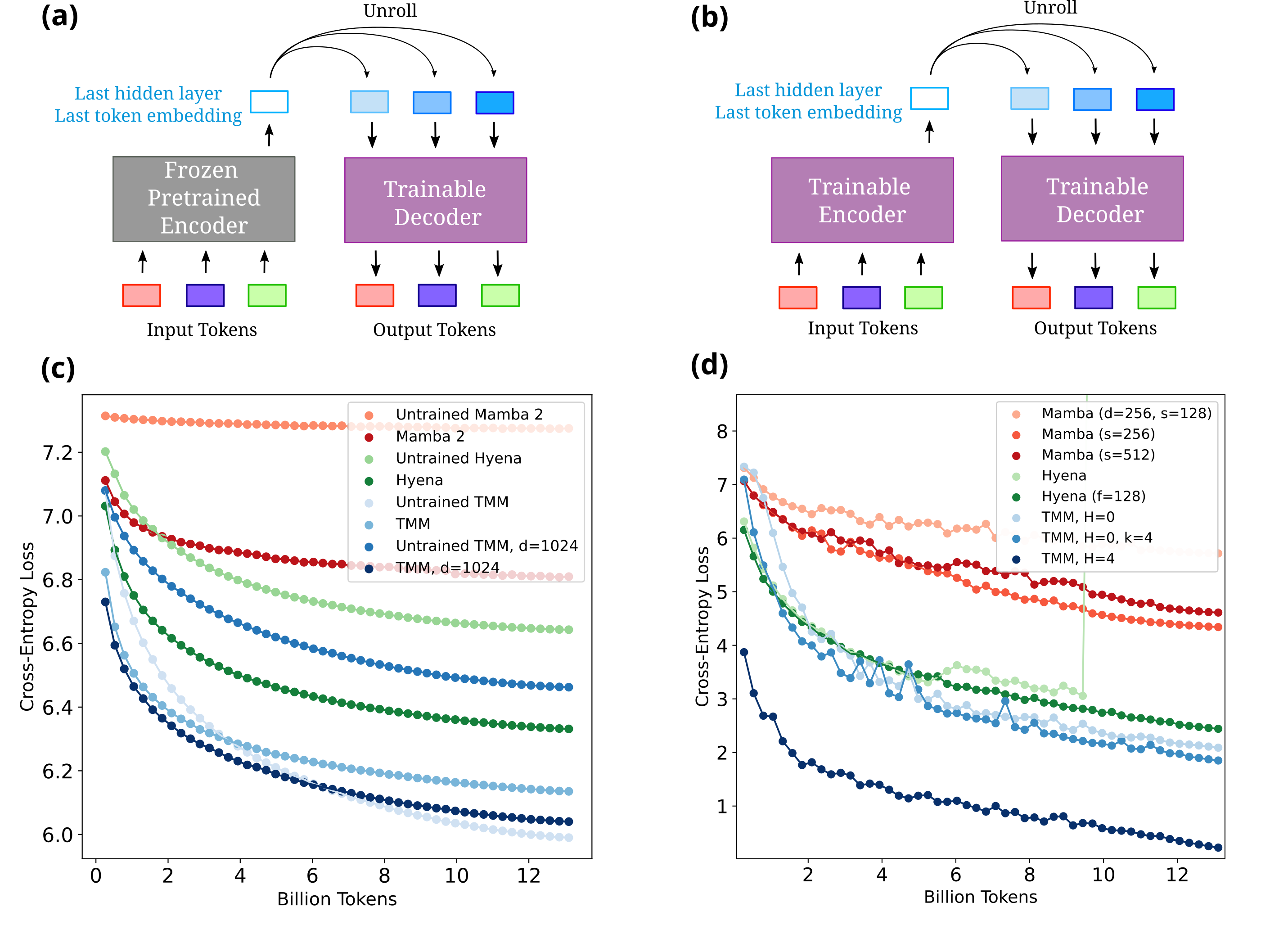}
        \caption{Information Retention and Capacity (a) Informational retention experimental design, (b) Informational Capacity experimental design, (c) Retention Results, (d) Capacity Results. $f=128$ indicates a filter order of 128 instead of 64 as used elsewhere for Hyena. All models $d_m=512, n_l=16$ unless otherwise noted.}
        \label{fig4}
    \end{figure}

    Information retention is an important aspect of all sub-quadratic complexity models because lookup of previous states is almost by definition restricted in some way. One measure of input information retention is the amount of information present in a model's last token, last hidden layer. In our experimental setting, a model is first pretrained for next token prediction (or initialized from scratch and not trained), frozen, and a decoder is initialized and trained to use the encoder's last token, last hidden layer embedding to regenerate the input sequence.  For all experiments, this embedding is `unrolled' (meaning a sliding window is projected) to form embedding given to the decoder as detailed in \citep{badger2025knowlimitsentropyestimation}. As shown in Figure \ref{fig4}, information retention increases for most models after causal training, with TMMs consistently retaining more input information than either Mamba or Hyena models.

    We next sought to understand the informational capacity of these models, which we define as the upper limit (more precisely the supremum) of informational retention in an architecture. What we test here is the ability of a model to retain input information if it is specifically trained to do so, and test this by initializing untrained encoders and decoders of a given architecture and training the model to regenerate the input. We find that TMMs exhibit larger informational retention capacity than either Hyena or Mamba models of equivalent $d_m$, despite requiring 1/2 to 1/8th the compute to train (Figure \ref{fig4}).
    
    We then investigated whether the trainable per-token parameters in Toeplitz operators were necessary for the high informational capacity of TMMs, and we find that this is indeed the case: trainable Toeplitz TMMs exhibited far higher informational capacity than TMMs without trainable Toeplitz layers, and the latter with trainable head projections are capable of much more information capacity than those without head projections (Figure \ref{figs2}).

    \subsection{Copy Ability} \label{copysection}

    Previously we observed information present in the last hidden layer of the last token of a given model, but the information present in all layers for many tokens remains to be determined. A functional test of this is the ability of a model to copy a certain number of tokens in its input to its output. We formulated a test of copying ability training after \citep{jelassi2024repeatmetransformersbetter}, where the rate at which various models learned to copy a sequence of tokens was found to correlate with pretrained models' copy ability, lookup accuracy, and other information retention characteristics. We train models to copy 512 tokens from sequences sampled from a natural language corpus (FineWeb) which were duplicated without special token delimiters, and models were evaluated on their accuracy in copying exactly 512 tokens from an in-distribution hold-out evaluation dataset in one forward pass.
        
\begin{figure}[h]
        \centering
        \includegraphics[width=0.95\textwidth]{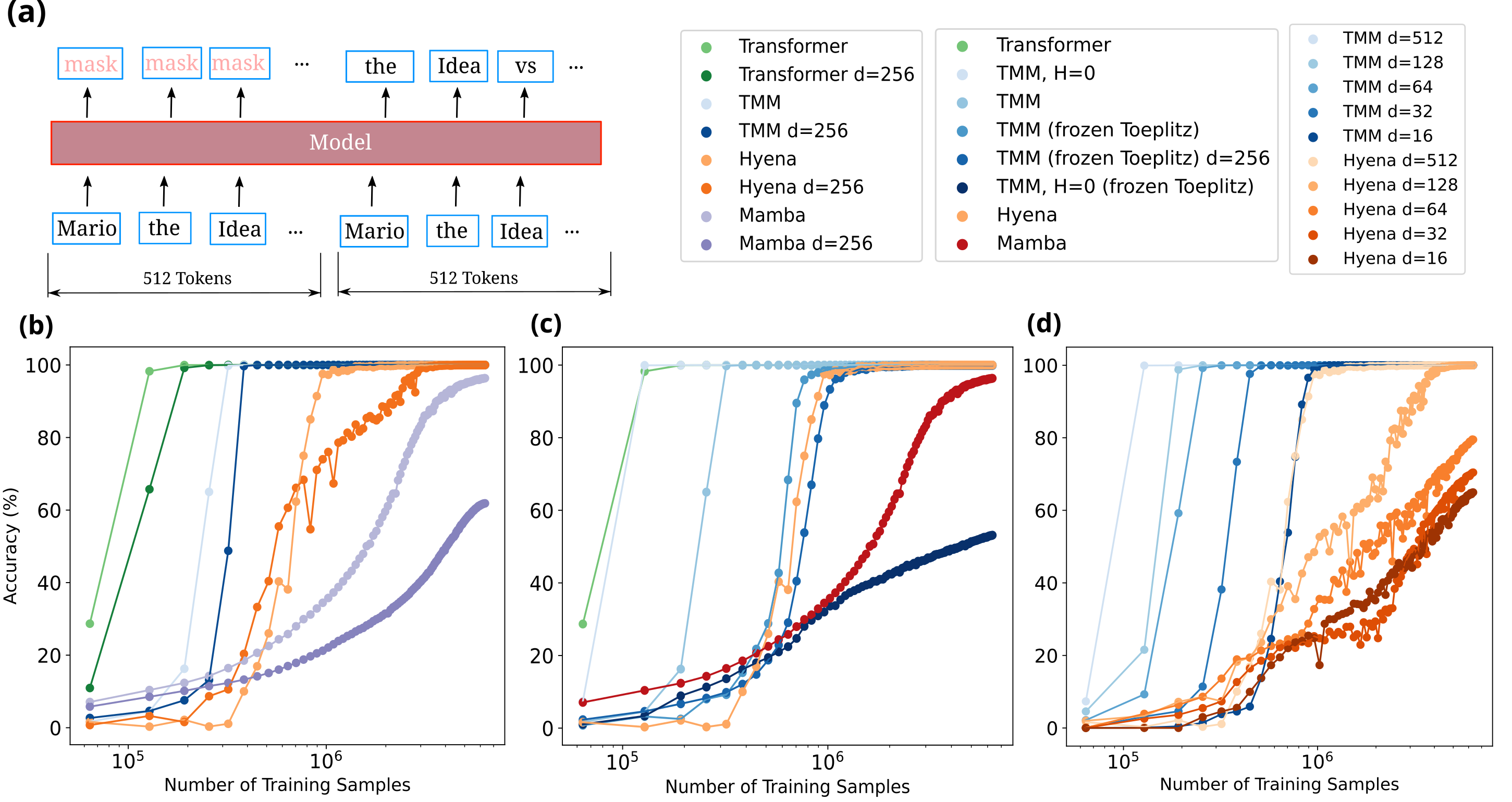}
        \caption{Copy task training efficiency. (a) Copy task depiction and Legends: left for (b), center (c), right (d). All models are $d_m=512$ and TMMs are non-headed unless otherwise noted. }
        \label{fig5}
    \end{figure}

    We find that Mamba models exhibit relatively poor copy task training as previously reported \citep{jelassi2024repeatmetransformersbetter}, that Hyena models learn with substantially fewer samples than Mamba, and that TMMs approach or even exceed the copy training efficiency of Transformers (Figure \ref{fig5} (a), and (b)). Investigating the relationship of copy training and model size scaling, we find that the gap between Hyena models and TMMs is substantial: a TMM with a $d_m$ of $1/32$x that of a Hyena model achieve nearly identical copy efficiency.
    
    Trainable Toeplitz parameters are necessary for the TMM's copy ability to approach or exceed that of the Transformer, and these parameters imbue far more copying efficiency than head projections: the non-headed frozen Toeplitz TMM is incapable of accurately copying after the entire training run, whereas the non-headed TMM with trainable Toeplitz layers exhibits the highest copy training efficiency of all tested models (Figure \ref{fig5}).

    \subsection{Benchmark Evaluation}

\begin{figure}
      \begin{minipage}[b]{.46\linewidth}
        \centering
        \includegraphics[width=0.99\linewidth]{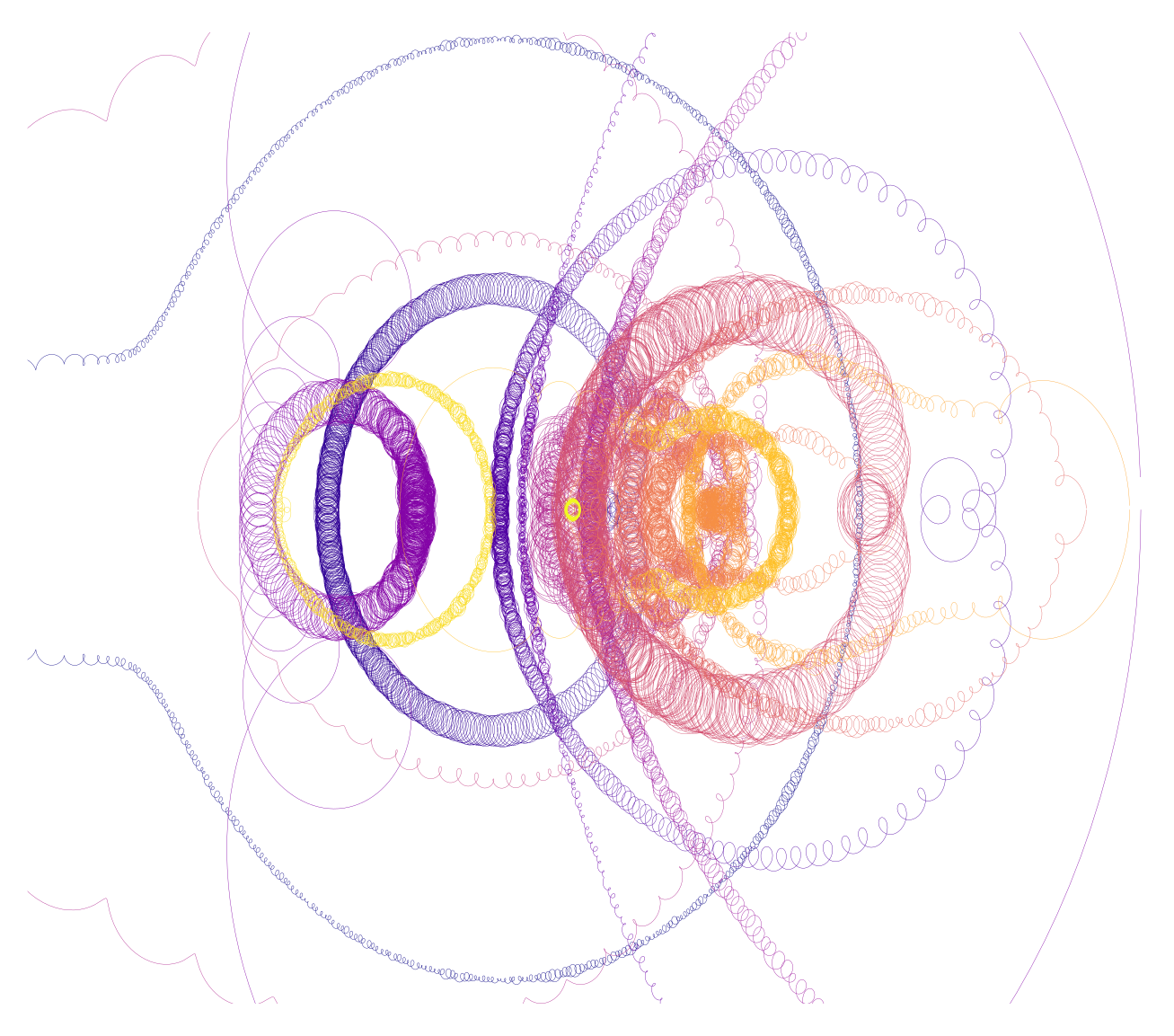}
      \end{minipage}\hfill
      \begin{minipage}[b]{.54\linewidth}
        \centering
        \renewcommand{\arraystretch}{1.1}
        \footnotesize
        \begin{tabular}{l c | c c} 
         \hline
          \textbf{IR/ICL Benchmarks}  & \textbf{Transformer} & \textbf{Hyena} & \textbf{TMM} \\
          SQuAD (exact) & - & 0.2 & \textbf{2.1} \\
          SQuAD v2 & & & \\
          \; \;  Has Exact & 5.1 & 0.0 & \textbf{1.7} \\
          \; \; F1 & - & 36.3 & \textbf{40.9} \\
          IFEval (strict prompt) & 12.2 & 7.8 & \textbf{8.7} \\
          LongBench & & & \\
          \; \; \scriptsize Few-Shot Learning & 4.5 & 2.3 & \textbf{2.5} \\
          \; \; \scriptsize Document QA & 1.5 & \textbf{2.3} & 2.0 \\
          \; \; \scriptsize Summarization & 7.0 & 5.7 & 5.7 \\
          \; \; \scriptsize Information Retrieval & 2.9 & 2.8 & \textbf{3.1} \\
          SWDE (contains) & 21.1 & 3.5 & \textbf{3.7} \\
          xWinoGrad (en) & 57.7 & 56.3 & \textbf{57.6} \\
         \hline 
         \textbf{Other Benchmarks}  & \textbf{Transformer} & \textbf{Hyena} & \textbf{TMM} \\
          ARC easy & 49.3 & 49.6 & 49.7 \\
          GLUE & 40.3 & 41.9 & \textbf{42.7} \\
          Lambada-OpenAI & 23.6 & \textbf{7.2} & 4.0 \\
          HellaSwag & 29.6 & 28.6 & 28.7 \\
          \hline
        \end{tabular}
        \vspace{5px}
      \end{minipage}
      \caption{Left: Toeplitz Symbols for trained next-token-prediction TMMs, color-coded per layer. Right: In-Context Learning, Information Retrieval, and general language understanding benchmarks (note that this Hyena model reaches a slightly lower loss than this TMM). All exact/strict matches where applicable, 0-shot.}
      \label{fig1}
    \end{figure}

\begin{figure}[h]
        \centering
        \includegraphics[width=0.45\textwidth]{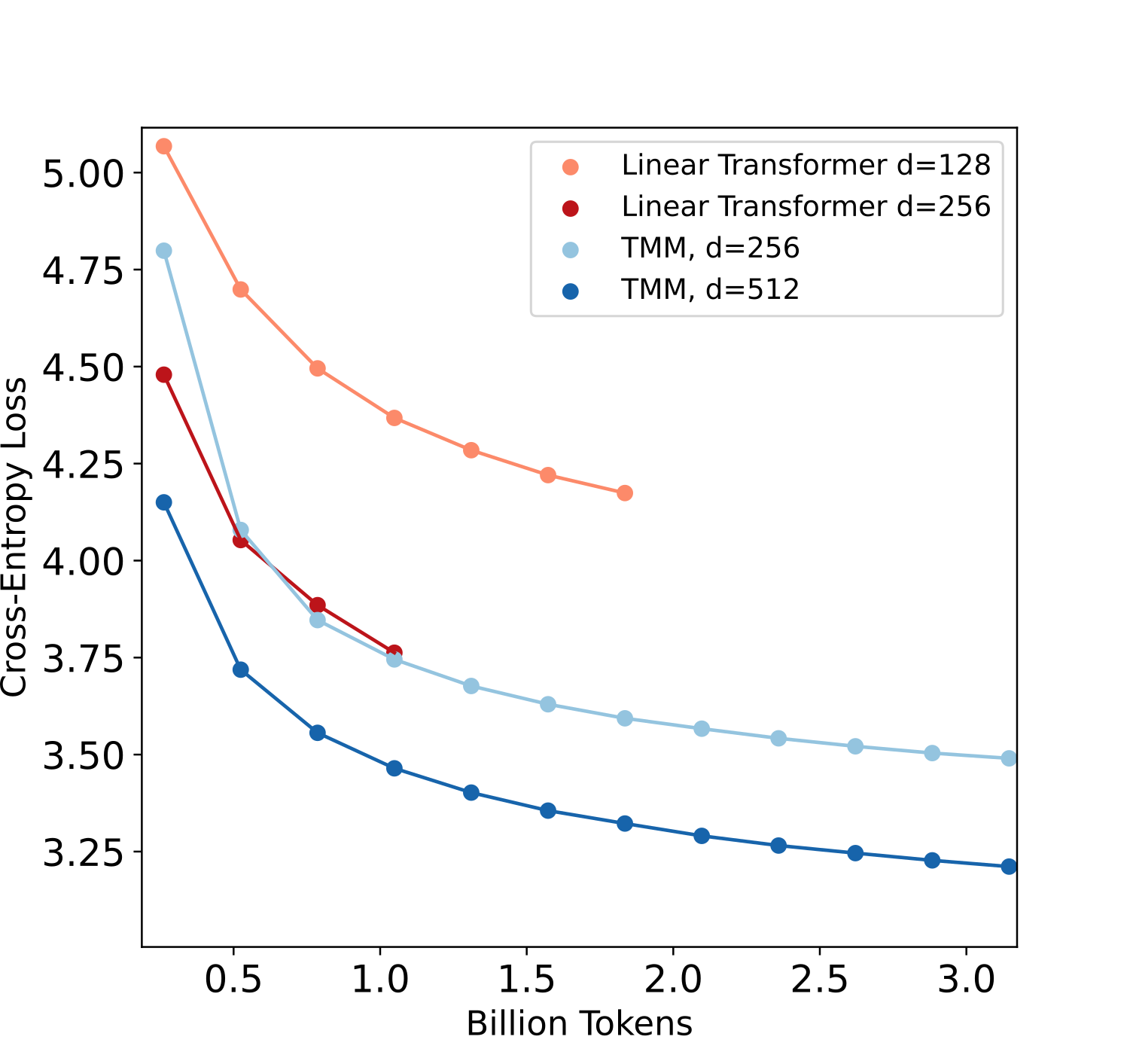}
        \caption{Causal Linear Transformers are inefficient to train and experience catastrophic numerical instabilities early in training. Linear Transformer plots halt where loss infinities are observed, and note that these instabilities are only observed for causal Linear Transformers.}
        \label{figs0}
    \end{figure}

\begin{figure}[h]
        \centering
        \includegraphics[width=0.95\textwidth]{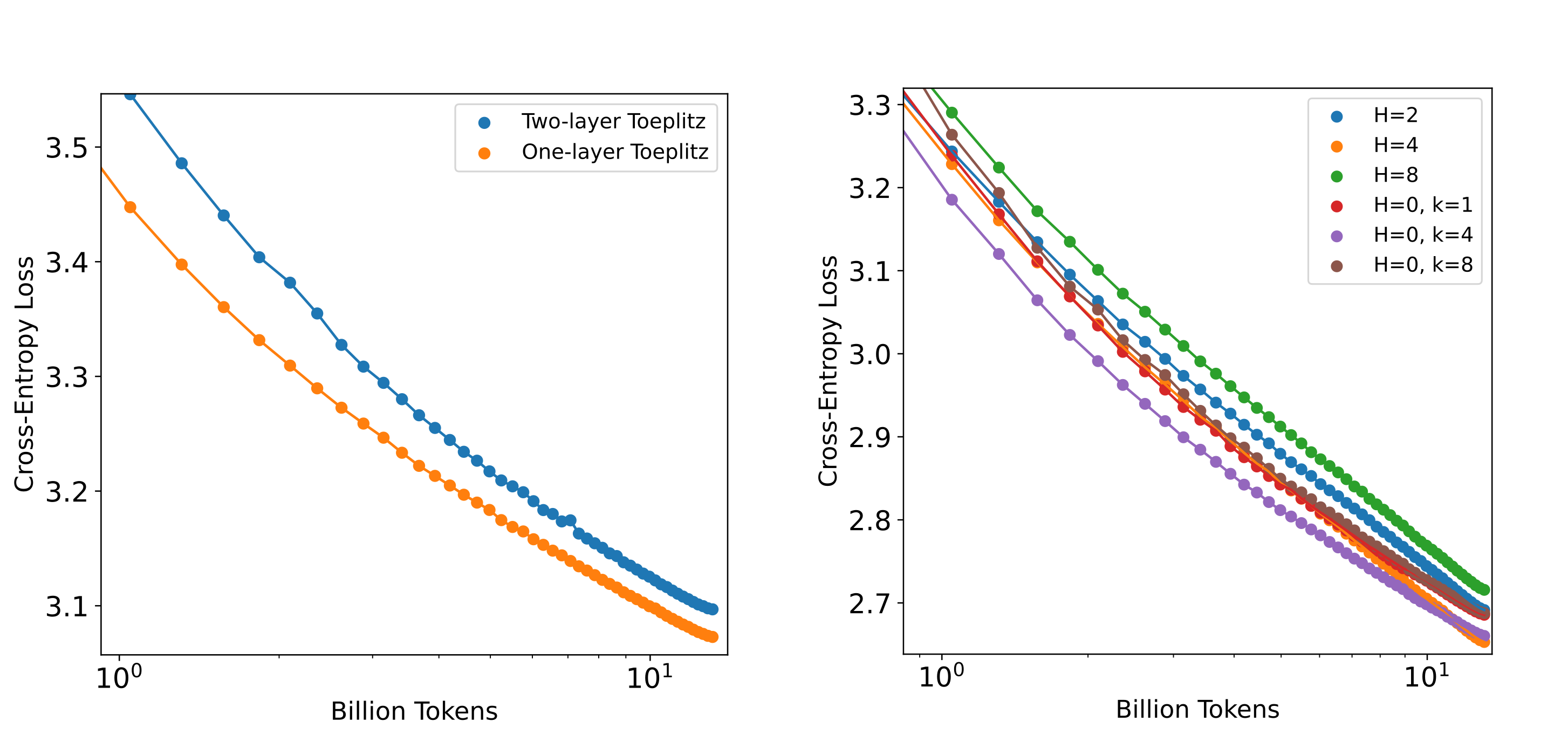}
        \caption{TMM architecture and FineWeb training efficiencies. Left, single Toeplitz versus two-layer (separated by GeLU nonlinearity) token mixing training efficiency, $d_m=512$ and MLP hidden layers are $d_m*2$ rather than $d_m*4$. Right, $d_m=1024, n_{ctx}=512, n_l=16$}
        \label{figs1}
    \end{figure}

\begin{figure}[h]
        \centering
        \includegraphics[width=0.95\textwidth]{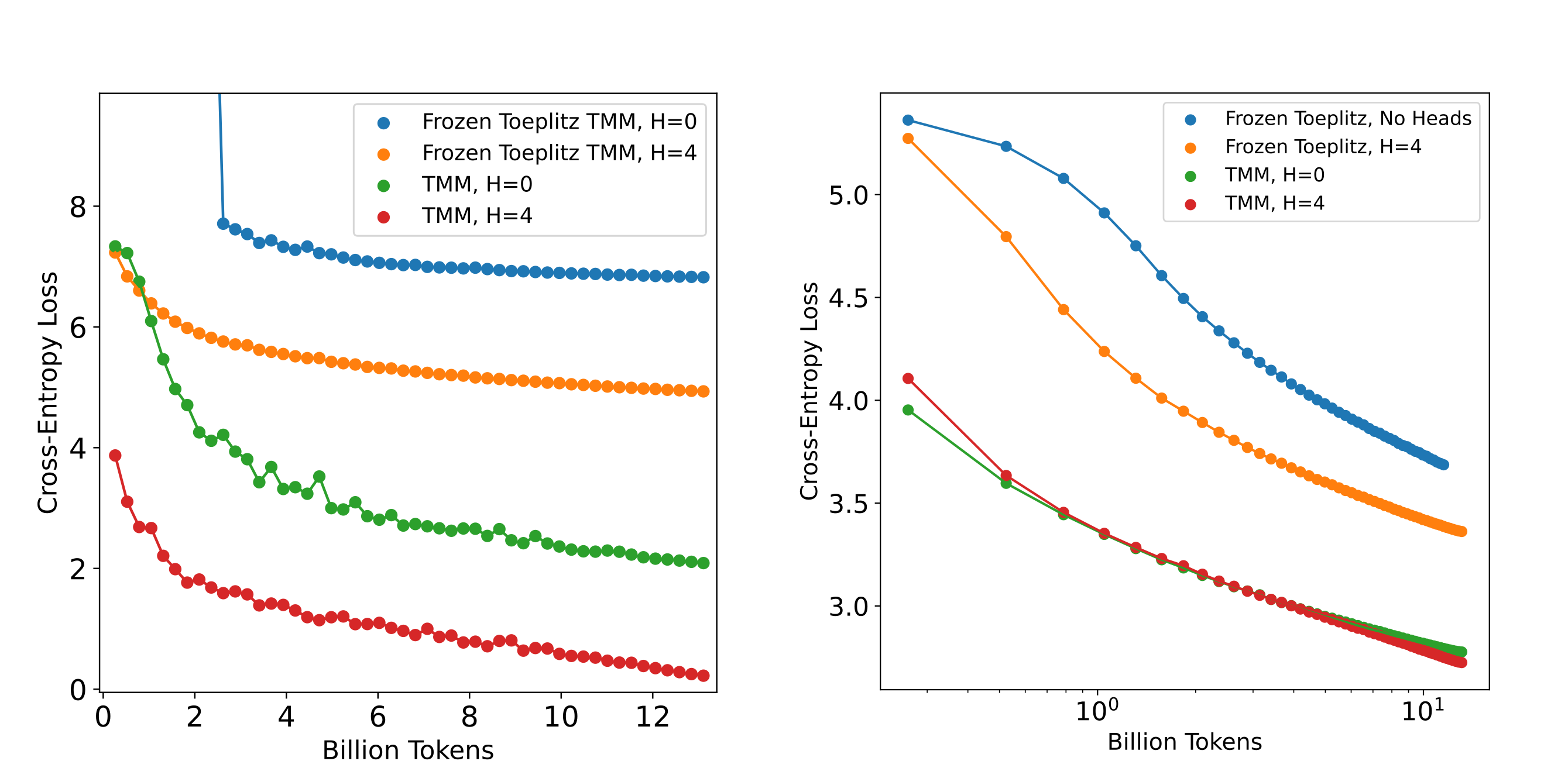}
        \caption{Trainable per-token parameters and information capacity. Left: TMMS with trainable (unfrozen) Toeplitz layers are more efficient to train than TMMs with frozen Toeplitz layers, regardless of whether trainable heads are used (all $d_m=1024, n_{ctx}=1024$). Right: FineWeb training efficiency for these models mirrors information capacity.}
        \label{figs2}
    \end{figure}

\begin{figure}[h]
        \centering
        \includegraphics[width=0.45\textwidth]{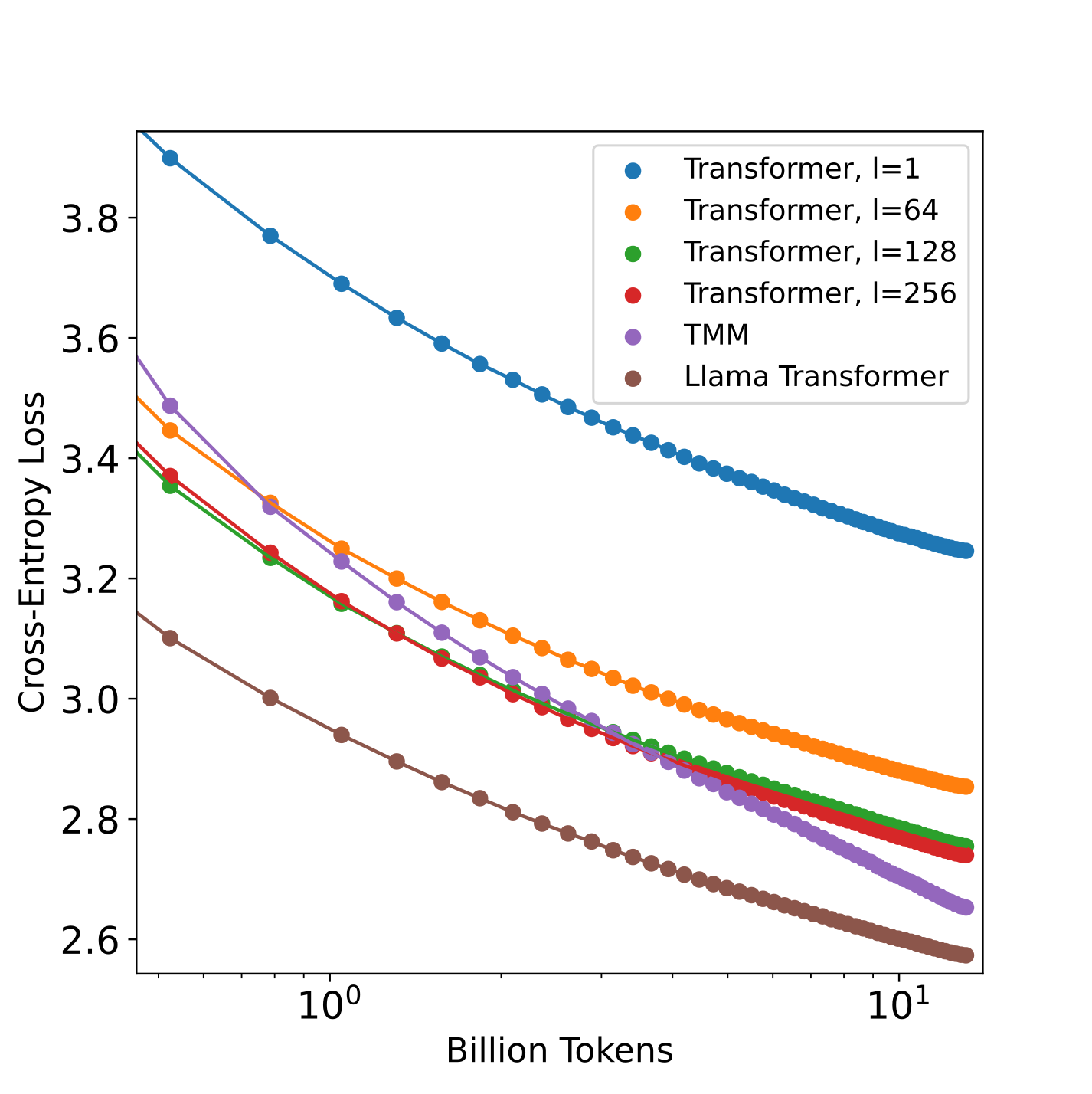}
        \caption{Toeplitz Mixers are more efficiently trainable than Windowed Attention Transformer models. All models $H=4, n_l=16, n_{ctx}=512$, Transformers are $d_m=512$ and TMMs are $d_m=1024$ to remain compute-matched.}
        \label{figs3}
    \end{figure}

    The higher information retention and capacity of TMMs compared to other low-complexity models suggests that these models are better suited to language tasks that require retention of input information in some way. We focus on two particular tasks in order to assess this hypothesis: information retrieval and in-context-learning, both of which require a model to maintain a more or less precise representation of the input sequence. We adapt the popular EleutherAI LM evaluation harness \citep{eval-harness} for use with our non-Transformer architectures by providing compatible model loading and forward pass modules and integrating these into a fully featured template. For this we benchmark on SQuAD \citep{rajpurkar2016squad100000questionsmachine}, SQuAD 2 \citep{rajpurkar-etal-2018-know}, Longbench \citep{bai2024longbench}, IFEval \citep{zhou2023instructionfollowingevaluationlargelanguage}, SWDE \citep{arora2024swde}, and xWinoGrad \citep{tikhonov2021heads} and focus on the performance of TMMs and Hyena models trained on equivalent data with equivalent compute. We compare these results to benchmarks of these models that do not require such precise input information handling: ARC-Easy \citep{Clark2018arc}, GLUE \citep{wang-etal-2018-glue}, Lambada (OpenAI processed) \citep{paperno2016lambadadatasetwordprediction}, and HellaSwag \citep{zellers2019hellaswagmachinereallyfinish}. As shown in Figure \ref{fig1}, TMMs typically outperform Hyena models for information retrieval and in-context learning but exhibit near parity performance on general language tasks, consistent with the superior information retention in trained TMMs compared to Hyena. 

    \section{Analysis and Discussion}

    We note that there are substantial differences in the Toeplitz weights of models that are trained to retain as much input information as possible compared to those that are trained to predict next tokens. Figure \ref{figs5} depicts all Toeplitz matrix weights for these models, and here it is clear that models trained for next token prediction learn a strong recency bias in terms of weight magnitude per token position, but that this recency bias is completely absent from models trained to maximize information retention. It is therefore unsurprising that Hyena models (with some amount of recency bias effectively mandated in the architecture) train to predict next words quickly but struggle to retain input information, as this bias is likely detrimental to information retention.

    \subsection{Inference Benefits of TMMs}

    TMMs exhibit lower time and space complexity than Transformers at inference prefill, allowing for the processing of longer sequences in a fixed time for a given set of hardware limitations (vRAM, FLOPs etc.). It is also clear that TMMs are more readily parallelizable at prefill than Transformers due to the linear rather than quadratic space complexity. Parallelizable prefill is useful for applications where many users attempt to inference on large contexts simultaneously, but is also beneficial for batched inference when a single correct output is desired from a given input, and where the input sequence may be modified in various ways to increase the likelihood of a correct output. In practical terms, this means that if one wishes to maximize the likelihood of generating an output that fulfills a certain requirement, the superior prefill parallelization properties of TMMs compared to Transformers means that it is more feasible to generate a large number of sequences from long inputs and thus maximize the likelihood of finding a good output.
    
    \subsection{Toeplitz Operator Symbols and Indices}

    Toeplitz matrices of trained models may be examined through the lens of operator algebras in order to answer the following questions: are Toeplitz matrices from trained models invertible, and if they are not what kind of conditions would need to be made in order to make them invertible? In particular, we seek to better understand the Toeplitz mixer operations for models that are specifically trained for invertibility over their input space (specifically autoencoders and copy models) versus those that we have shown to be non-invertible over their inputs (causal models, untrained models). It should be noted that TMMs (like Transformers, Masked Mixers, and most other architectures) are not strictly invertible across all possible inputs as they contain abundant non-invertible transformations making the meaning of invertibility of one specific layer somewhat dubious; this is our motivation to explore the Fredholm indices rather than invertibility only, as the latter can be computed by simply finding the matrix rank via SVD decomposition.

    The definition of the Fredholm index of a linear operator $T$ is the difference between the dimension of the kernel of $T$ and its cokernel, and is not to be confused with the kernel and cokernel of the Toeplitz matrix itself (these are equal as $T \in \Bbb R^{N \times N}$ is square) and given in Equation \ref{eq9} for convenience.

    \begin{equation}
    F_i = \mathrm{dim}(\mathrm{ker}(T)) - \mathrm{dim} (\mathrm{coker}(T))
    \label{eq9}
    \end{equation}

    We use the Toeplitz index theorem to measure the indices of Toeplitz matrices of our models. This theorem states that the Fredholm index of a Toeplitz matrix is equivalent to the negative winding number of the symbol of $T$, defined for a Toeplitz matrix represented as a vector with indices $(b_{-n}, b_{-n-1}, ..., b_0, b_1, ...b_n)$ given points $z \in \Bbb C$ on the unit circle of the complex plane as follows:

    \begin{equation}
    I(T) = -\sigma(T) = - \sum_n b_n z^{-n}
    \label{eq10}
    \end{equation}

\begin{figure}[h]
        \centering
        \includegraphics[width=0.85\textwidth]{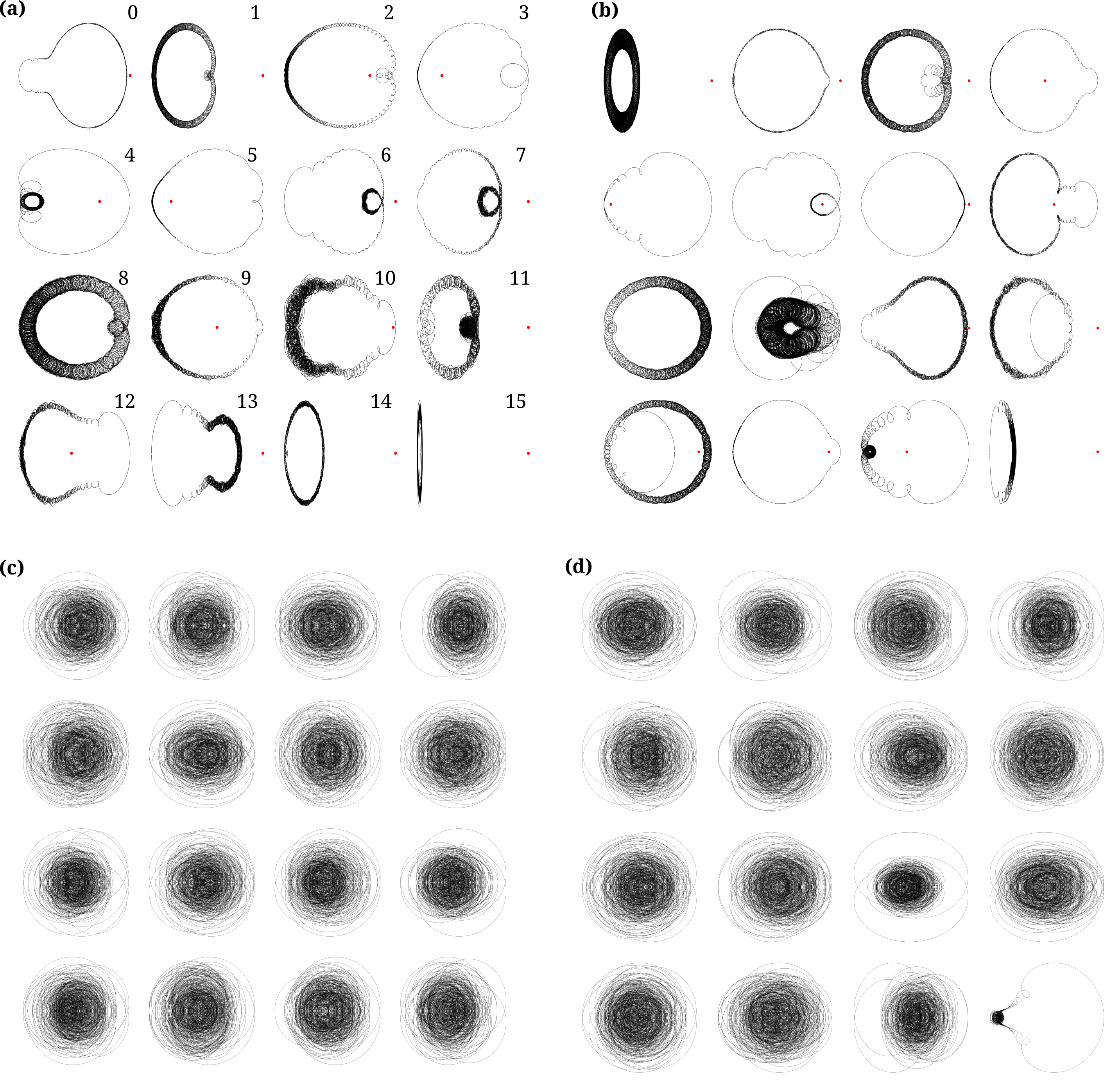}
        \caption{Trained TMM model Toeplitz Symbols per model layer in the complex plane. (a) d=1024 causal model, (b) d=245 causal model, (c) d=512 autoencoder encoder, (d) d=512 autoencoder decoder. All non-headed models trained on the FineWeb. Red dots denote the origin.}
        \label{figs4}
    \end{figure}

    We plot symbols of Toeplitz matrices of various models for in Figure \ref{figs4} for clarity, and include a plot of all symbols of one causal-trained CLM superimposed on each other in Figure \ref{fig1} to illustrate their alignment.
    
    Winding number computation was modified to account for our use of Toeplitz matrices as follows: by the Toeplitz index theorem the operator's index is negative the winding number for left multiplication, but as we right-multiply the masked Toeplitz matrices in our models we effectively swaps the kernel and cokernel in Equation \ref{eq9} such that the index of these operators is equal to the winding number. Indices for all Toeplitz layers in various models are given in Table \ref{table1}.

    \begin{center}
    \begin{table}[H]
    \scriptsize
    \begin{center}
    \renewcommand{\arraystretch}{1.1}
    \begin{tabular}{l c c c c c c c c c c c c c c c c c} 
     \hline
      Layer  & 0 & 1 & 2 & 3 & 4 & 5 & 6 & 7 & 8 & 9 & 10 & 11 & 12 & 13 & 14 & 15 & \\
     \hline \hline
      Untrained & -480& -502& -511& -492& -516& -510& -517& -486& -498& -472& -513& -468& -494& -500& -489& -490 & \\
     \hline
      CLM &  &  &  &  &  &  &  &  &  &  &  &  &  &  &  &  &  \\
      d=1024 & 0 & 0 & -1 & -1 & -1 & -1 & 0 & 0 & -14 & -1 & -1 & 0 & -1 & 0 & 0 & 0 & \\
      d=256 & 0 & 0 & 0 & -1 & -1 & -2 & 0 & -1 & -33 & -14 & 0 & 0 & -1 & -1 & -1 & 0 &\\
     \hline 
      Copy & -406 & -400 & -417 & -412 & -428 & -440 & -438 & -417 & -457 & -449 & -471 & -452 & -455 & -430 & -469 & -399 & \\
      \hline 
      Auto  &  &  &  &  &  &  &  &  &  &  &  &  &  &  &  &  &  \\
      Encoder & -247& -232& -248& -249& -252& -245& -229& -240& -251& -236& -254& -244& -264& -252& -245 & -240 & \\
      Decoder & -267& -248& -261& -249& -243& -229& -255& -247& -269& -271& -257& -259& -279& -261& -231& 0 & \\
      \hline 
    \end{tabular}
    \end{center}
    \vspace{0.1cm}
    \caption{Fredholm Indices per Toeplitz mixer layer for various models, $n_{ctx}=512$ except for the copy model (which is $n_{ctx}=1024$) and all models are non-headed.}
    \label{table1}
    \end{table}
    \end{center}

    We draw a number of notable conclusions from these measurements: firstly that perfect information retention does not require invertible Toeplitz matrices as invertible operations are necessarily $I(T)=0$ (presumably this is because these models only need to be invertible over their input space, not over all possible inputs), secondly that models that exhibit larger information retention may contain larger (absolute) indices, and thirdly that there are remarkable similarities between indices of a given model or model type, with the notable exception of the last layer in the Autoencoder decoder compared to the other layers of that model.

\begin{figure}[h]
        \centering
        \includegraphics[width=0.85\textwidth]{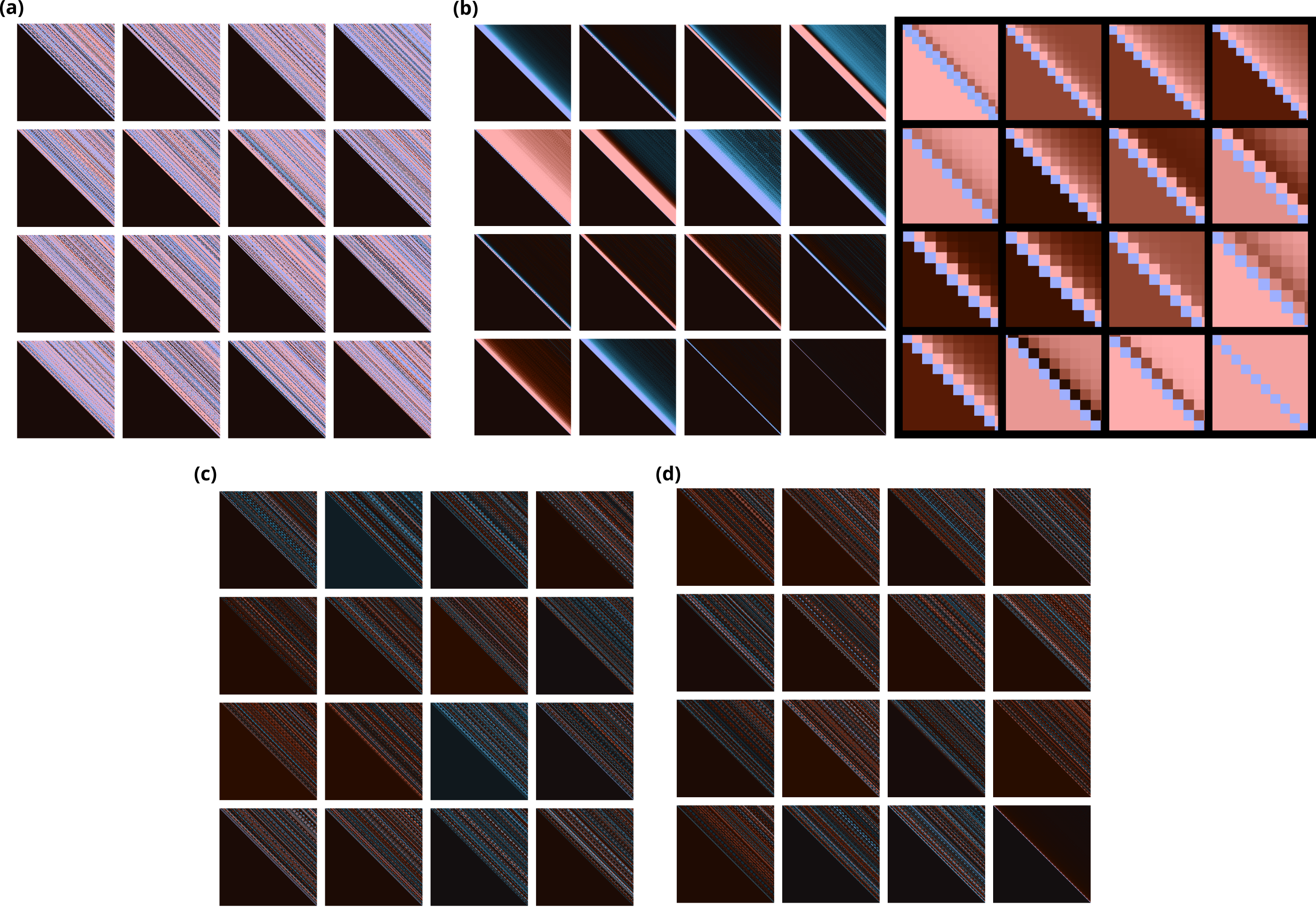}
        \caption{Toeplitz Weight Matrix Values. Blue is negative, Red positive. (a) Untrained, (b) Causal trained: Left, all weights with range (-0.015, 0.015) and Right, weights near the main diagonal with automatic scaling, (c) autoencoder encoder, (d) autoencoder decoder}
        \label{figs5}
    \end{figure}

    The most common non-zero index of causal models is -1, which is the index of the shift operator that maps $(a, b, c, ...) \to (0, a, b, ...)$ and may be expressed as a Toeplitz matrix with 1s one index above the main diagonal, 0 elsewhere for right multiplication. Consulting the magnitudes of causal Toeplitz weight matrices near the main diagonal (Figure \ref{figs5} (b) right), we can see that indeed these layers qualitatively resemble this operator such that we can assume that information is typically shifted by one or a few tokens per layer in these models. As these layers are sequentially arranged in the TMMs, the total token shift in the model is the sum of the token shift per layer. This suggests that deeper TMMs trained for causal language modeling would exhibit superior information retention, and likely explains the benefit of training deeper Hyena models as well \citep{poli2023hyenahierarchylargerconvolutional}.
    
    \section{Limitations}

    There are a number of limitations to this work, the most significant of which are enumerated here. Firstly, this work is limited in scale: models are typically between 20 to 300 million parameters, trained on 13 or 26 billion tokens. We observe superior information access characteristics of TMMs compared to other sub-quadratic models at this scale, but there is no \textit{a priori} guarantee that this finding is applicable to models trained on many orders of magnitude more compute.

    A notable functional limitation of TMMs is that they remain $\mathcal O(n^2)$ complexity for autoregressive generation once the first token has been computed. Practically speaking this means that at inference, on can expect for a TMM to exhibit far lower time to first token for very long prompts but that once the first token is generated, subsequent tokens will require approximately the same amount of time and memory to generate as would be required for a Transformer.
    
    The models investigated are furthermore limited by the number of configurations tested, which is out of necessity as there are an exponential number of options for architectural choices. We focus on similarly-sized models and primarily change the model width $d_m$, but it remains possible that different configurations or hypoerparameters would yield better or worse training efficiency or information processing characteristics.
    
    \section{Reproducibility}
    
    Code for this work is found on Github: https://github.com/ethan-w-roland/ToeplitzMixers. All training experiments in this paper may be launched from that repo, and require relatively modest computational resources to run. For benchmark evaluations, we modified the EleutherAI \texttt{lm-evaluation-harness} for use with the appropriate models; that code may be found at https://github.com/blbadger/lm-evaluation-harness.
    
    \section{Conclusion}
    
    TMMs provide a simple, attention-free path to competitive training efficiency and fast decoding, with stronger information preservation than other attention-free baselines.
    
    \bibliographystyle{unsrtnat}
    \bibliography{references}  

\begin{thebibliography}{37}
\providecommand{\natexlab}[1]{#1}
\providecommand{\url}[1]{\texttt{#1}}
\expandafter\ifx\csname urlstyle\endcsname\relax
  \providecommand{\doi}[1]{doi: #1}\else
  \providecommand{\doi}{doi: \begingroup \urlstyle{rm}\Url}\fi

\bibitem[Vaswani et~al.(2023)Vaswani, Shazeer, Parmar, Uszkoreit, Jones, Gomez, Kaiser, and Polosukhin]{vaswani2023attentionneed}
Ashish Vaswani, Noam Shazeer, Niki Parmar, Jakob Uszkoreit, Llion Jones, Aidan~N. Gomez, Lukasz Kaiser, and Illia Polosukhin.
\newblock Attention is all you need.
\newblock 2023.
\newblock URL \url{https://arxiv.org/abs/1706.03762}.

\bibitem[Radford et~al.(2019)Radford, Wu, Child, Luan, Amodei, Sutskever, et~al.]{radford2019language}
Alec Radford, Jeffrey Wu, Rewon Child, David Luan, Dario Amodei, Ilya Sutskever, et~al.
\newblock Language models are unsupervised multitask learners.
\newblock \emph{OpenAI blog}, 1\penalty0 (8):\penalty0 9, 2019.

\bibitem[Shen et~al.(2018)Shen, Zhang, Zhao, Yi, and Li]{shen2019efficient}
Zhuoran Shen, Mingyuan Zhang, Haiyu Zhao, Shuai Yi, and Hongsheng Li.
\newblock Efficient attention: Attention with linear complexities.
\newblock \emph{CoRR}, abs/1812.01243, 2018.
\newblock URL \url{http://arxiv.org/abs/1812.01243}.

\bibitem[Katharopoulos et~al.(2020)Katharopoulos, Vyas, Pappas, and Fleuret]{katharopoulos2020}
Angelos Katharopoulos, Apoorv Vyas, Nikolaos Pappas, and François Fleuret.
\newblock Transformers are rnns: Fast autoregressive transformers with linear attention, 2020.
\newblock URL \url{https://arxiv.org/abs/2006.16236}.

\bibitem[Poli et~al.(2023)Poli, Massaroli, Nguyen, Fu, Dao, Baccus, Bengio, Ermon, and Ré]{poli2023hyenahierarchylargerconvolutional}
Michael Poli, Stefano Massaroli, Eric Nguyen, Daniel~Y. Fu, Tri Dao, Stephen Baccus, Yoshua Bengio, Stefano Ermon, and Christopher Ré.
\newblock Hyena hierarchy: Towards larger convolutional language models, 2023.
\newblock URL \url{https://arxiv.org/abs/2302.10866}.

\bibitem[Gu and Dao(2024)]{gu2024mamba}
Albert Gu and Tri Dao.
\newblock Mamba: Linear-time sequence modeling with selective state spaces.
\newblock In \emph{First conference on language modeling}, 2024.

\bibitem[Dao and Gu(2024)]{dao2024transformersssmsgeneralizedmodels}
Tri Dao and Albert Gu.
\newblock Transformers are ssms: Generalized models and efficient algorithms through structured state space duality.
\newblock 2024.
\newblock URL \url{https://arxiv.org/abs/2405.21060}.

\bibitem[Jelassi et~al.(2024)Jelassi, Brandfonbrener, Kakade, and Malach]{jelassi2024repeatmetransformersbetter}
Samy Jelassi, David Brandfonbrener, Sham~M. Kakade, and Eran Malach.
\newblock Repeat after me: Transformers are better than state space models at copying, 2024.
\newblock URL \url{https://arxiv.org/abs/2402.01032}.

\bibitem[Waleffe et~al.(2024)Waleffe, Byeon, Riach, Norick, Korthikanti, Dao, Gu, Hatamizadeh, Singh, Narayanan, Kulshreshtha, Singh, Casper, Kautz, Shoeybi, and Catanzaro]{waleffe2024empiricalstudymambabasedlanguage}
Roger Waleffe, Wonmin Byeon, Duncan Riach, Brandon Norick, Vijay Korthikanti, Tri Dao, Albert Gu, Ali Hatamizadeh, Sudhakar Singh, Deepak Narayanan, Garvit Kulshreshtha, Vartika Singh, Jared Casper, Jan Kautz, Mohammad Shoeybi, and Bryan Catanzaro.
\newblock An empirical study of mamba-based language models.
\newblock 2024.
\newblock URL \url{https://arxiv.org/abs/2406.07887}.

\bibitem[Fu et~al.(2023{\natexlab{a}})Fu, Dao, Saab, Thomas, Rudra, and Ré]{fu2023hungryhungryhipposlanguage}
Daniel~Y. Fu, Tri Dao, Khaled~K. Saab, Armin~W. Thomas, Atri Rudra, and Christopher Ré.
\newblock Hungry hungry hippos: Towards language modeling with state space models, 2023{\natexlab{a}}.
\newblock URL \url{https://arxiv.org/abs/2212.14052}.

\bibitem[Fu et~al.(2023{\natexlab{b}})Fu, Arora, Grogan, Johnson, Eyuboglu, Thomas, Spector, Poli, Rudra, and Ré]{fu2023monarchmixersimplesubquadratic}
Daniel~Y. Fu, Simran Arora, Jessica Grogan, Isys Johnson, Sabri Eyuboglu, Armin~W. Thomas, Benjamin Spector, Michael Poli, Atri Rudra, and Christopher Ré.
\newblock Monarch mixer: A simple sub-quadratic gemm-based architecture, 2023{\natexlab{b}}.
\newblock URL \url{https://arxiv.org/abs/2310.12109}.

\bibitem[Badger(2025)]{badger2025maskedmixerslanguagegeneration}
Benjamin~L. Badger.
\newblock Masked mixers for language generation and retrieval.
\newblock 2025.
\newblock URL \url{https://arxiv.org/abs/2409.01482}.

\bibitem[Melas-Kyriazi(2021)]{melaskyriazi2021needattentionstackfeedforward}
Luke Melas-Kyriazi.
\newblock Do you even need attention? a stack of feed-forward layers does surprisingly well on imagenet, 2021.
\newblock URL \url{https://arxiv.org/abs/2105.02723}.

\bibitem[Tolstikhin et~al.(2021)Tolstikhin, Houlsby, Kolesnikov, Beyer, Zhai, Unterthiner, Yung, Steiner, Keysers, Uszkoreit, Lucic, and Dosovitskiy]{tolstikhin2021mlpmixerallmlparchitecturevision}
Ilya Tolstikhin, Neil Houlsby, Alexander Kolesnikov, Lucas Beyer, Xiaohua Zhai, Thomas Unterthiner, Jessica Yung, Andreas Steiner, Daniel Keysers, Jakob Uszkoreit, Mario Lucic, and Alexey Dosovitskiy.
\newblock Mlp-mixer: An all-mlp architecture for vision, 2021.
\newblock URL \url{https://arxiv.org/abs/2105.01601}.

\bibitem[Virtanen et~al.(2020)Virtanen, Gommers, Oliphant, Haberland, Reddy, Cournapeau, Burovski, Peterson, Weckesser, Bright, et~al.]{virtanen2020scipy}
Pauli Virtanen, Ralf Gommers, Travis~E Oliphant, Matt Haberland, Tyler Reddy, David Cournapeau, Evgeni Burovski, Pearu Peterson, Warren Weckesser, Jonathan Bright, et~al.
\newblock Scipy 1.0: fundamental algorithms for scientific computing in python.
\newblock \emph{Nature methods}, 17\penalty0 (3):\penalty0 261--272, 2020.

\bibitem[Liu et~al.(2022)Liu, Jiao, and Lim]{liu2022ludecompositiontoeplitzdecomposition}
Yucong Liu, Simiao Jiao, and Lek-Heng Lim.
\newblock Lu decomposition and toeplitz decomposition of a neural network, 2022.
\newblock URL \url{https://arxiv.org/abs/2211.13935}.

\bibitem[Beltagy et~al.(2020)Beltagy, Peters, and Cohan]{beltagy2020longformerlongdocumenttransformer}
Iz~Beltagy, Matthew~E. Peters, and Arman Cohan.
\newblock Longformer: The long-document transformer, 2020.
\newblock URL \url{https://arxiv.org/abs/2004.05150}.

\bibitem[Kitaev et~al.(2020)Kitaev, Łukasz Kaiser, and Levskaya]{kitaev2020reformerefficienttransformer}
Nikita Kitaev, Łukasz Kaiser, and Anselm Levskaya.
\newblock Reformer: The efficient transformer, 2020.
\newblock URL \url{https://arxiv.org/abs/2001.04451}.

\bibitem[Hochreiter and Schmidhuber(1997)]{hochreiter1997long}
Sepp Hochreiter and J{\"u}rgen Schmidhuber.
\newblock Long short-term memory.
\newblock \emph{Neural computation}, 9\penalty0 (8):\penalty0 1735--1780, 1997.

\bibitem[Penedo et~al.(2024)Penedo, Kydlíček, allal, Lozhkov, Mitchell, Raffel, Werra, and Wolf]{penedo2024finewebdatasetsdecantingweb}
Guilherme Penedo, Hynek Kydlíček, Loubna~Ben allal, Anton Lozhkov, Margaret Mitchell, Colin Raffel, Leandro~Von Werra, and Thomas Wolf.
\newblock The fineweb datasets: Decanting the web for the finest text data at scale, 2024.
\newblock URL \url{https://arxiv.org/abs/2406.17557}.

\bibitem[Paszke et~al.(2019)Paszke, Gross, Massa, Lerer, Bradbury, Chanan, Killeen, Lin, Gimelshein, Antiga, Desmaison, Köpf, Yang, DeVito, Raison, Tejani, Chilamkurthy, Steiner, Fang, Bai, and Chintala]{paszke2019pytorchimperativestylehighperformance}
Adam Paszke, Sam Gross, Francisco Massa, Adam Lerer, James Bradbury, Gregory Chanan, Trevor Killeen, Zeming Lin, Natalia Gimelshein, Luca Antiga, Alban Desmaison, Andreas Köpf, Edward Yang, Zach DeVito, Martin Raison, Alykhan Tejani, Sasank Chilamkurthy, Benoit Steiner, Lu~Fang, Junjie Bai, and Soumith Chintala.
\newblock Pytorch: An imperative style, high-performance deep learning library, 2019.
\newblock URL \url{https://arxiv.org/abs/1912.01703}.

\bibitem[Wolf et~al.(2020)Wolf, Debut, Sanh, Chaumond, Delangue, Moi, Cistac, Rault, Louf, Funtowicz, Davison, Shleifer, von Platen, Ma, Jernite, Plu, Xu, Scao, Gugger, Drame, Lhoest, and Rush]{wolf-etal-2020-transformers}
Thomas Wolf, Lysandre Debut, Victor Sanh, Julien Chaumond, Clement Delangue, Anthony Moi, Pierric Cistac, Tim Rault, Rémi Louf, Morgan Funtowicz, Joe Davison, Sam Shleifer, Patrick von Platen, Clara Ma, Yacine Jernite, Julien Plu, Canwen Xu, Teven~Le Scao, Sylvain Gugger, Mariama Drame, Quentin Lhoest, and Alexander~M. Rush.
\newblock Transformers: State-of-the-art natural language processing.
\newblock In \emph{Proceedings of the 2020 Conference on Empirical Methods in Natural Language Processing: System Demonstrations}, pages 38--45, Online, October 2020. Association for Computational Linguistics.
\newblock URL \url{https://www.aclweb.org/anthology/2020.emnlp-demos.6}.

\bibitem[Lhoest et~al.(2021)Lhoest, Villanova~del Moral, Jernite, Thakur, von Platen, Patil, Chaumond, Drame, Plu, Tunstall, Davison, {\v{S}}a{\v{s}}ko, Chhablani, Malik, Brandeis, Le~Scao, Sanh, Xu, Patry, McMillan-Major, Schmid, Gugger, Delangue, Matussi{\`e}re, Debut, Bekman, Cistac, Goehringer, Mustar, Lagunas, Rush, and Wolf]{lhoest-etal-2021-datasets}
Quentin Lhoest, Albert Villanova~del Moral, Yacine Jernite, Abhishek Thakur, Patrick von Platen, Suraj Patil, Julien Chaumond, Mariama Drame, Julien Plu, Lewis Tunstall, Joe Davison, Mario {\v{S}}a{\v{s}}ko, Gunjan Chhablani, Bhavitvya Malik, Simon Brandeis, Teven Le~Scao, Victor Sanh, Canwen Xu, Nicolas Patry, Angelina McMillan-Major, Philipp Schmid, Sylvain Gugger, Cl{\'e}ment Delangue, Th{\'e}o Matussi{\`e}re, Lysandre Debut, Stas Bekman, Pierric Cistac, Thibault Goehringer, Victor Mustar, Fran{\c{c}}ois Lagunas, Alexander Rush, and Thomas Wolf.
\newblock Datasets: A community library for natural language processing.
\newblock In \emph{Proceedings of the 2021 Conference on Empirical Methods in Natural Language Processing: System Demonstrations}, pages 175--184, Online and Punta Cana, Dominican Republic, November 2021. Association for Computational Linguistics.
\newblock URL \url{https://aclanthology.org/2021.emnlp-demo.21}.

\bibitem[Loshchilov and Hutter(2019)]{loshchilov2019decoupledweightdecayregularization}
Ilya Loshchilov and Frank Hutter.
\newblock Decoupled weight decay regularization, 2019.
\newblock URL \url{https://arxiv.org/abs/1711.05101}.

\bibitem[OLMo et~al.(2025)OLMo, Walsh, Soldaini, Groeneveld, Lo, Arora, Bhagia, Gu, Huang, Jordan, Lambert, Schwenk, Tafjord, Anderson, Atkinson, Brahman, Clark, Dasigi, Dziri, Ettinger, Guerquin, Heineman, Ivison, Koh, Liu, Malik, Merrill, Miranda, Morrison, Murray, Nam, Poznanski, Pyatkin, Rangapur, Schmitz, Skjonsberg, Wadden, Wilhelm, Wilson, Zettlemoyer, Farhadi, Smith, and Hajishirzi]{olmo20252olmo2furious}
Team OLMo, Pete Walsh, Luca Soldaini, Dirk Groeneveld, Kyle Lo, Shane Arora, Akshita Bhagia, Yuling Gu, Shengyi Huang, Matt Jordan, Nathan Lambert, Dustin Schwenk, Oyvind Tafjord, Taira Anderson, David Atkinson, Faeze Brahman, Christopher Clark, Pradeep Dasigi, Nouha Dziri, Allyson Ettinger, Michal Guerquin, David Heineman, Hamish Ivison, Pang~Wei Koh, Jiacheng Liu, Saumya Malik, William Merrill, Lester James~V. Miranda, Jacob Morrison, Tyler Murray, Crystal Nam, Jake Poznanski, Valentina Pyatkin, Aman Rangapur, Michael Schmitz, Sam Skjonsberg, David Wadden, Christopher Wilhelm, Michael Wilson, Luke Zettlemoyer, Ali Farhadi, Noah~A. Smith, and Hannaneh Hajishirzi.
\newblock 2 olmo 2 furious, 2025.
\newblock URL \url{https://arxiv.org/abs/2501.00656}.

\bibitem[He et~al.(2015)He, Zhang, Ren, and Sun]{he2015delvingdeeprectifierssurpassing}
Kaiming He, Xiangyu Zhang, Shaoqing Ren, and Jian Sun.
\newblock Delving deep into rectifiers: Surpassing human-level performance on imagenet classification, 2015.
\newblock URL \url{https://arxiv.org/abs/1502.01852}.

\bibitem[Gao et~al.(2024)Gao, Tow, Abbasi, Biderman, Black, DiPofi, Foster, Golding, Hsu, Le~Noac'h, Li, McDonell, Muennighoff, Ociepa, Phang, Reynolds, Schoelkopf, Skowron, Sutawika, Tang, Thite, Wang, Wang, and Zou]{eval-harness}
Leo Gao, Jonathan Tow, Baber Abbasi, Stella Biderman, Sid Black, Anthony DiPofi, Charles Foster, Laurence Golding, Jeffrey Hsu, Alain Le~Noac'h, Haonan Li, Kyle McDonell, Niklas Muennighoff, Chris Ociepa, Jason Phang, Laria Reynolds, Hailey Schoelkopf, Aviya Skowron, Lintang Sutawika, Eric Tang, Anish Thite, Ben Wang, Kevin Wang, and Andy Zou.
\newblock The language model evaluation harness, 07 2024.
\newblock URL \url{https://zenodo.org/records/12608602}.

\bibitem[Rajpurkar et~al.(2016)Rajpurkar, Zhang, Lopyrev, and Liang]{rajpurkar2016squad100000questionsmachine}
Pranav Rajpurkar, Jian Zhang, Konstantin Lopyrev, and Percy Liang.
\newblock Squad: 100,000+ questions for machine comprehension of text, 2016.
\newblock URL \url{https://arxiv.org/abs/1606.05250}.

\bibitem[Rajpurkar et~al.(2018)Rajpurkar, Jia, and Liang]{rajpurkar-etal-2018-know}
Pranav Rajpurkar, Robin Jia, and Percy Liang.
\newblock Know what you don{'}t know: Unanswerable questions for {SQ}u{AD}.
\newblock In Iryna Gurevych and Yusuke Miyao, editors, \emph{Proceedings of the 56th Annual Meeting of the Association for Computational Linguistics (Volume 2: Short Papers)}, pages 784--789, Melbourne, Australia, July 2018. Association for Computational Linguistics.
\newblock \doi{10.18653/v1/P18-2124}.
\newblock URL \url{https://aclanthology.org/P18-2124}.

\bibitem[Bai et~al.(2024)Bai, Lv, Zhang, Lyu, Tang, Huang, Du, Liu, Zeng, Hou, Dong, Tang, and Li]{bai2024longbench}
Yushi Bai, Xin Lv, Jiajie Zhang, Hongchang Lyu, Jiankai Tang, Zhidian Huang, Zhengxiao Du, Xiao Liu, Aohan Zeng, Lei Hou, Yuxiao Dong, Jie Tang, and Juanzi Li.
\newblock {L}ong{B}ench: A bilingual, multitask benchmark for long context understanding.
\newblock In \emph{Proceedings of the 62nd Annual Meeting of the Association for Computational Linguistics (Volume 1: Long Papers)}, pages 3119--3137, Bangkok, Thailand, August 2024. Association for Computational Linguistics.
\newblock \doi{10.18653/v1/2024.acl-long.172}.
\newblock URL \url{https://aclanthology.org/2024.acl-long.172}.

\bibitem[Zhou et~al.(2023)Zhou, Lu, Mishra, Brahma, Basu, Luan, Zhou, and Hou]{zhou2023instructionfollowingevaluationlargelanguage}
Jeffrey Zhou, Tianjian Lu, Swaroop Mishra, Siddhartha Brahma, Sujoy Basu, Yi~Luan, Denny Zhou, and Le~Hou.
\newblock Instruction-following evaluation for large language models, 2023.
\newblock URL \url{https://arxiv.org/abs/2311.07911}.

\bibitem[Arora et~al.(2024)Arora, Eyuboglu, Zhang, Timalsina, Alberti, Zinsley, Zou, Rudra, and Ré]{arora2024swde}
Simran Arora, Sabri Eyuboglu, Michael Zhang, Aman Timalsina, Silas Alberti, Dylan Zinsley, James Zou, Atri Rudra, and Christopher Ré.
\newblock Simple linear attention language models balance the recall-throughput tradeoff, 2024.

\bibitem[Tikhonov and Ryabinin(2021)]{tikhonov2021heads}
Alexey Tikhonov and Max Ryabinin.
\newblock It's all in the heads: Using attention heads as a baseline for cross-lingual transfer in commonsense reasoning, 2021.

\bibitem[Clark et~al.(2018)Clark, Cowhey, Etzioni, Khot, Sabharwal, Schoenick, and Tafjord]{Clark2018arc}
Peter Clark, Isaac Cowhey, Oren Etzioni, Tushar Khot, Ashish Sabharwal, Carissa Schoenick, and Oyvind Tafjord.
\newblock Think you have solved question answering? try arc, the ai2 reasoning challenge.
\newblock \emph{ArXiv}, abs/1803.05457, 2018.

\bibitem[Wang et~al.(2018)Wang, Singh, Michael, Hill, Levy, and Bowman]{wang-etal-2018-glue}
Alex Wang, Amanpreet Singh, Julian Michael, Felix Hill, Omer Levy, and Samuel Bowman.
\newblock {GLUE}: A multi-task benchmark and analysis platform for natural language understanding.
\newblock In \emph{Proceedings of the 2018 {EMNLP} Workshop {B}lackbox{NLP}: Analyzing and Interpreting Neural Networks for {NLP}}, pages 353--355, Brussels, Belgium, November 2018. Association for Computational Linguistics.
\newblock \doi{10.18653/v1/W18-5446}.
\newblock URL \url{https://aclanthology.org/W18-5446}.

\bibitem[Paperno et~al.(2016)Paperno, Kruszewski, Lazaridou, Pham, Bernardi, Pezzelle, Baroni, Boleda, and Fernández]{paperno2016lambadadatasetwordprediction}
Denis Paperno, Germán Kruszewski, Angeliki Lazaridou, Quan~Ngoc Pham, Raffaella Bernardi, Sandro Pezzelle, Marco Baroni, Gemma Boleda, and Raquel Fernández.
\newblock The lambada dataset: Word prediction requiring a broad discourse context, 2016.
\newblock URL \url{https://arxiv.org/abs/1606.06031}.

\bibitem[Zellers et~al.(2019)Zellers, Holtzman, Bisk, Farhadi, and Choi]{zellers2019hellaswagmachinereallyfinish}
Rowan Zellers, Ari Holtzman, Yonatan Bisk, Ali Farhadi, and Yejin Choi.
\newblock Hellaswag: Can a machine really finish your sentence?, 2019.
\newblock URL \url{https://arxiv.org/abs/1905.07830}.

\end{thebibliography}
    
    \beginsupplement
    \section{Appendix}

    \subsection{Training Resource Requirements} \label{throughputsection}

    Here provide tables of typical throughputs during training for models of various hidden dimensions $d_m$, benchmarked on a 2x H100 server. Most training was performed using a 4x V100 server, and as such models were scaled to be able to fit in memory and be trainable in a reasonable amount of time on that machine ($d_m=1024, \;n_l=16, \;b=128$ for TMMs for instance.) Training throughputs for these models using that machine were proportionally similar for the H100s, with the exception of Mamba models which are very slow to train on V100s. All measurements are made without \texttt{torch.compile}, although Mamba 2 models use compiled modules: for all Mamba models, we use the fast Mamba kernels available at https://github.com/state-spaces/mamba. 

    \begin{center}
    \begin{table}[H]
    \small
    \begin{center}
    \renewcommand{\arraystretch}{1.1}
    \begin{tabular}{||l c c ||} 
     \hline
      dm & Throughput (it/s) & Memory per GPU (MB) \\
     \hline \hline
      128 & 8.38 & 11299 \\
      256 & 6.71 & 14595 \\
      512 & 4.56 & 23959 \\
      1024 & 2.47 & 42831 \\
      2048 & 1.06 & 85755 \\
      \hline 
    \end{tabular}
    \end{center}
    \vspace{0.1cm}
    \caption{Hyena Throughput and Memory requirements, 2x H100 ($b=128$, $n_{ctx}=512$) }
    \label{tables1}
    \end{table}
    \end{center}

    \begin{center}
    \begin{table}[H]
    \small
    \begin{center}
    \renewcommand{\arraystretch}{1.1}
    \begin{tabular}{||l c c ||} 
     \hline
      dm & Throughput (it/s) & Memory per GPU (MB) \\
     \hline \hline
      128 & 13.57 & 9923 \\
      256 & 11.76 & 12217 \\
      512 & 8.81 & 16721 \\
      1024 & 4.85 & 28253 \\
      2048 & 1.85 & 53589 \\
      \hline 
    \end{tabular}
    \end{center}
    \vspace{0.1cm}
    \caption{TMM (H=4) Throughput and Memory requirements, 2x H100 ($b=128$, $n_{ctx}=512$) }
    \label{tables2}
    \end{table}
    \end{center}

    \begin{center}
    \begin{table}[H]
    \small
    \begin{center}
    \renewcommand{\arraystretch}{1.1}
    \begin{tabular}{||l c c ||} 
     \hline
      dm & Throughput (it/s) & Memory per GPU (MB) \\
     \hline \hline
      128 & 14.5 & 7431 \\
      256 & 12.61 & 9399 \\
      512 & 9.30 & 16721 \\
      1024 & 5.25 & 22741 \\
      2048 & 2.07 & 45285 \\
      \hline 
    \end{tabular}
    \end{center}
    \vspace{0.1cm}
    \caption{TMM (No Heads) Throughput and Memory requirements, 2x H100 ($b=128$, $n_{ctx}=512$) }
    \label{tables3}
    \end{table}
    \end{center}

\begin{center}
    \begin{table}[H]
    \small
    \begin{center}
    \renewcommand{\arraystretch}{1.1}
    \begin{tabular}{||l c c ||} 
     \hline
      dm & Throughput (it/s) & Memory per GPU (MB) \\
     \hline \hline
      128 & 8.02 & 14329 \\
      256 & 6.10 & 21713 \\
      512 & 3.84 & 34491 \\
      1024 & 1.90 & 63295 \\
      2048 & 0.71* & 170022* \\
      \hline 
    \end{tabular}
    \end{center}
    \vspace{0.1cm}
    \caption{Mamba Throughput and Memory requirements, 2x H100 ($b=128$, $n_{ctx}=512$). * with gradient accumulation.}
    \label{tables4}
    \end{table}
    \end{center}

    \subsection{Information}

\end{document}